\pdfoutput=1

\documentclass[11pt]{article}

\usepackage[]{ACL2023}

\usepackage{times}
\usepackage{latexsym}

\usepackage[T1]{fontenc}

\usepackage[utf8]{inputenc}

\usepackage{microtype}

\usepackage{inconsolata}
\usepackage{booktabs}
\usepackage{multirow}
\usepackage{colortbl}  
\usepackage{xcolor}
\usepackage{array}   
\usepackage{subfigure}
\usepackage[graphicx]{realboxes}
\usepackage{tikz}
\usepackage{amsmath, amssymb}
\usepackage{adjustbox}
\usepackage{xspace}
\usepackage{hyperref}

\usepackage{bbding}
\usepackage{soul} 
\usepackage{color}
\usepackage{arydshln}
\usepackage{amsthm,amsmath,amssymb}
 
\usepackage{mathrsfs}
\usepackage{pifont}
\usepackage{circledsteps}
\usepackage{fleqn}

\title{OOP: Object-Oriented Programming Evaluation Benchmark\\ for Large Language Models}
\author{
    Paper ID: 3791
}

\author{Shuai Wang\textsuperscript{\rm 1}\space\space
Liang Ding\textsuperscript{\rm 2}\space\space
Li Shen\textsuperscript{\rm 3}\space\space
Yong Luo\textsuperscript{\rm 1}\space\space
Bo Du\textsuperscript{\rm 1}\space\space 
Dacheng Tao\textsuperscript{\rm 2}\\
    \textsuperscript{\rm 1}Wuhan University\space\space
    \textsuperscript{\rm 2}The University of Sydney\space\space 
    \textsuperscript{\rm 3}JD Explore Academy \\
    {\tt\small wangshuai123@whu.edu.cn},\space\space
    {\tt\small liangding.liam@gmail.com}
}

\begin{document}
\maketitle
\begin{abstract}
Advancing automated programming necessitates robust and comprehensive code generation benchmarks, yet current evaluation frameworks largely neglect \textbf{o}bject-\textbf{o}riented \textbf{p}rogramming (\textbf{OOP}) in favour of functional programming (FP), e.g., HumanEval and MBPP. To address this, \ding{182} our study introduces \textbf{a pioneering OOP-focused benchmark}, featuring 431 Python programs that encompass essential OOP concepts and features like classes and encapsulation methods. \ding{183} We propose \textbf{a novel evaluation metric, \textit{pass@$o$}}, tailored for OOP, enhancing traditional \textit{pass@$k$} metric. \ding{184} Our evaluation of $23$ leading large language models (LLMs), including both general and code-specialized models, \textbf{reveals three key insights}: 1) \textit{pass@$o$} offers a more relevant and comprehensive assessment for OOP code generation; 2) Despite excelling in FP, code-specialized LLMs like WizardCoder lag in OOP compared to models like ChatGPT; 3) The poor performance of all advanced LLMs on our OOP benchmark highlights a critical need for improvements in this field. Our benchmark and scripts are publicly released at: \url{https://github.com/alphadl/OOP-eval}.
\end{abstract}

\section{Introduction}
Large language models (LLMs,~\citealp[]{ouyang2022training,touvron2023llama}), consisting of billions or even trillions of parameters' Transformer blocks~\cite{vaswani2017attention}, have emerged like mushrooms after the rain, especially since the emergence of ChatGPT\footnote{\url{https://chat.openai.com}}.
In comparison to small models, LLMs exhibit stronger generalization and reasoning capabilities~\cite{wei2022emergent}. Currently, LLMs are playing a crucial role in various tasks, e.g., code generation~\cite{chen2021evaluating,li2022competition,roziere2023code}, language understanding~\cite{zhong2023chat}, human-computer interaction~\cite{tolomei2023prompt,moslem-etal-2023-adaptive}, and translation~\cite{Peng2023ChatGPT4MT,Lu2023EAPrompt}.
\begin{figure}[!t]
    \centering
    \includegraphics[scale=0.37]{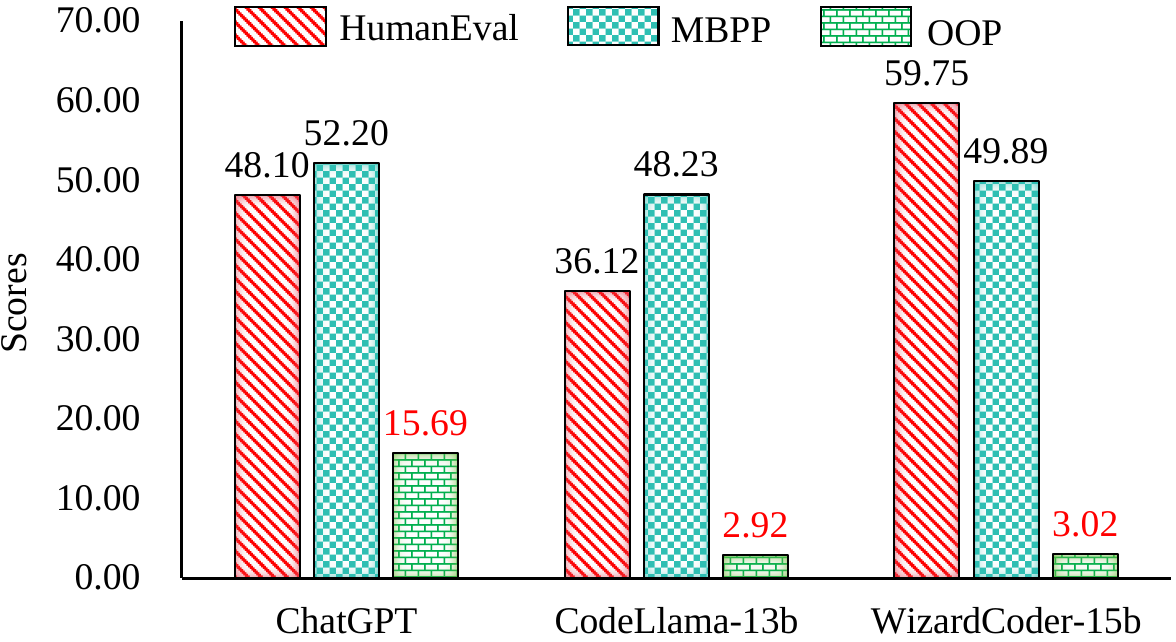}
    \caption{\textbf{The performance comparison of widely-used code language models on functional programming (FP) and object-oriented programming (OOP) code generation benchmarks}, in terms of \textit{pass@$1$} scores. We see that all models perform relatively well on FP benchmarks, i.e., Humaneval~\cite{chen2021evaluating} and MBPP~\cite{austin2021program}, while exhibiting poor performance on our OOP benchmark.}
    \label{fig:compcare_motivate}
\end{figure}

\begin{table*}[!ht]
  \centering
  \begin{adjustbox}{width=\textwidth}
    \begin{tabular}{lcccc}
    \toprule
    \textbf{Benchmark}  & \textbf{Number} & \textbf{NL}    & \textbf{PL}    & \textbf{Task Type} \\
    \midrule
    HumanEval \cite{chen2021evaluating} & $164$   & en    & Python & Function Programming \\
    MBPP \cite{austin2021program} & $974$   & en    & Python & Function Programming \\
    APPS \cite{hendrycks2021measuring} & $5000$  & en    & Python & Function Programming \\
    CodeContests \cite{li2022competition} & $165$   & en    & Multi & Function Programming \\
    MultiPL-MBPP \cite{cassano2023multipl} & $974^*$ & Multi & Multi & Function Programming \\
    HumanEval-X \cite{zheng2023codegeex} & $164^*$ & en    & Multi & Function Programming \\
    MultiPL-HumanEval \cite{cassano2023multipl} & $164^*$ & en    & Multi & Function Programming \\
    MTPB \cite{nijkamp2022codegen} & $115$   & en    & Python & Function Programming \\
    ODEX \cite{wang2022execution} & $945$   & Multi & Python & Function Programming \\
    PandasEval \cite{zan2022cert} & $101$   & en    & Python & Function Programming \\
    BIG-Bench \cite{srivastava2022beyond} & $32$    & en    & Python & Function Programming \\
    CodeApex$^\star$ \cite{fu2023codeapex} & $476^*$ & zh\&en & C++   & Function Programming \\
    \rowcolor[rgb]{ .906,  .902,  .902} OOP (Our)& $431$   & en    & Python & Object-Oriented Programming \\
    \bottomrule
    \end{tabular}%
    \end{adjustbox}
    \caption{\textbf{Overview of existing code evaluation benchmarks}. (``NL'' denotes natural language describing the problem or requirements; ``PL'' represents the generated programming language; ``en'' and ``zh'' denote English and Chinese, respectively, and ``Multi'' means containing multiple NLs or PLs; ``$^*$'' indicates the number of samples for each language; ``$\star$'' means that in CodeApex, we only considered code generation tasks.)}
  \label{tab:existing_benchmarks}%
\end{table*}%
The process of code generation entails crafting code in a suitable programming language from natural language descriptions of problems or requirements, aiming to effectively solve the problems or fulfill the requirements.
Given that hiring professional programmers to write code consumes a significant amount of human and material resources, the importance of automated programming becomes particularly evident.
Currently, the question of how to use the rising LLMs to generate more accurate automated programming codes based on problems or requirements stated by actual natural language has become an important research topic~\cite{liu2023your,zhong2023study}. In the research process, code generation evaluation is crucial. Code generation evaluation not only needs to objectively and impartially reflect the current performance of LLMs in programming but also should disclose the shortcomings in LLM programming to further enhance its potential.

\textbf{[Importance of OOP]} According to the November programming language rankings by TIOBE~\footnote{\url{https://www.tiobe.com/tiobe-index/}}, four out of the top five programming languages are OOP, which reflects the importance of OOP languages. OOP is centred on designing code around data or objects rather than organizing it based on functionality and logic~\cite{stroustrup1988object,stefik1985object}. OOP focuses more on the programming paradigm of class and object~\cite{wegner1990concepts}. Functions are commonly referred to as methods in OOP.



\textbf{[Motivation]} However, existing code generation evaluation benchmarks primarily focus on the evaluation of FP, and lack the evaluation of relevant concepts and features of OOP, e.g., class, inheritance, encapsulation methods, etc. If using existing benchmarks in Table~\ref{tab:existing_benchmarks} for evaluation can only show the performance of LLMs in FP, it fails to reflect their potential in OOP, as illustrated in Figure~\ref{fig:compcare_motivate}.


\textbf{[OOP benchmark and metric]} Considering the limitations of current code generation evaluation FP benchmarks and the widespread use of the Python programming language, we propose the first OOP evaluation benchmark based on Python. OOP benchmark consists of $431$ Python programs, covering key concepts and features of OOP, including class, inheritance, encapsulation methods, etc. Furthermore, to prevent the issue where LLMs may not generate concepts and features of OOP, we have optimized the \textit{pass@$k$}~\cite{kulal2019spoc,chen2021evaluating} metric by matching key points in natural language with key points in the programming language, i.e., the class names and private function names, etc, for natural language requirements are matched with the class names and private function names, etc., in the programming language. Our main \textbf{contributions} are summarized as follows:
\begin{enumerate}
\item We construct and release the first OOP evaluation benchmark, which encompasses concepts and features of OOP, e.g., class, polymorphism, encapsulation methods, etc.
\item We devise a new metric \textit{pass@$o$} based on conventional pass@$k$, tailored for the OOP code generation task, by matching key points in natural language and programming language.
\item We extensively evaluated our OOP with $23$ advanced LLMs, demonstrating that \textit{i)} there is still significant room for improving the OOP tasks, \textit{ii)} our benchmark could serve as a robust and fair indicator that helps the community quantify LLMs' OOP performance.
\end{enumerate}

\section{Related work}
\paragraph{Code Evaluation Benchmark} In the early days of LLMs, researchers from Google and OpenAI launched artificial handwritten code evaluation benchmarks, namely MBPP~\cite{austin2021program} and HumanEval~\cite{chen2021evaluating}, respectively. MBPP and HumanEval are currently the mainstream code generation evaluation benchmarks, but both of them are based on the Python programming language.
Subsequently, MultiPL-MBPP~\cite{cassano2023multipl} and MultiPL-HumanEval~\cite{cassano2023multipl} expanded upon these two benchmarks by translating the Python programming language into eighteen other programming languages, e.g., Java, C++, PHP, etc, to evaluate the performance of LLMs across others programming languages. Additionally, HumanEval-X~\cite{zheng2023codegeex} incorporated multiple test cases into the HumanEval benchmark. Apart from the extensions made to these two benchmarks, other benchmarks like CodeApex~\cite{fu2023codeapex} and ODEX~\cite{wang2022execution} exhibit distinctive features across different natural languages and task types. Unlike existing code evaluation benchmarks, our proposed OOP benchmark primarily focuses on the concepts and features of OOP, e.g., class, inheritance, etc. These works are summarized in Table~\ref{tab:existing_benchmarks}.

\paragraph{Code Evaluation Metrics} Existing evaluation metrics can be broadly categorized into two types: dynamic evaluation metrics and static evaluation metrics. Dynamic evaluation metrics evaluate the executability of generated codes by using test cases, with \textit{pass@$k$}~\cite{kulal2019spoc,chen2021evaluating} serving as the primary representative. The calculation process for \textit{pass@$k$} is shown in Appendix~\ref{sec:appendix_related_work}. Additionally, this category of metrics includes \textit{$n$@$k$}~\cite{li2022competition}. Static evaluation metrics calculate BLUE~\cite{papineni2002bleu}, ROUGE~\cite{lin2004rouge}, Codescore~\cite{dong2023codescore} and CodeBLEU~\cite{ren2020codebleu} among manually written examples and generated programs. However, these code evaluation metrics do not specifically focus on evaluating the concepts and features of OOP. Therefore, we further optimized the \textit{pass@$k$} metric based on the evaluation benchmark for OOP.

\begin{figure}[!t]
    \centering
    \includegraphics[scale=0.65]{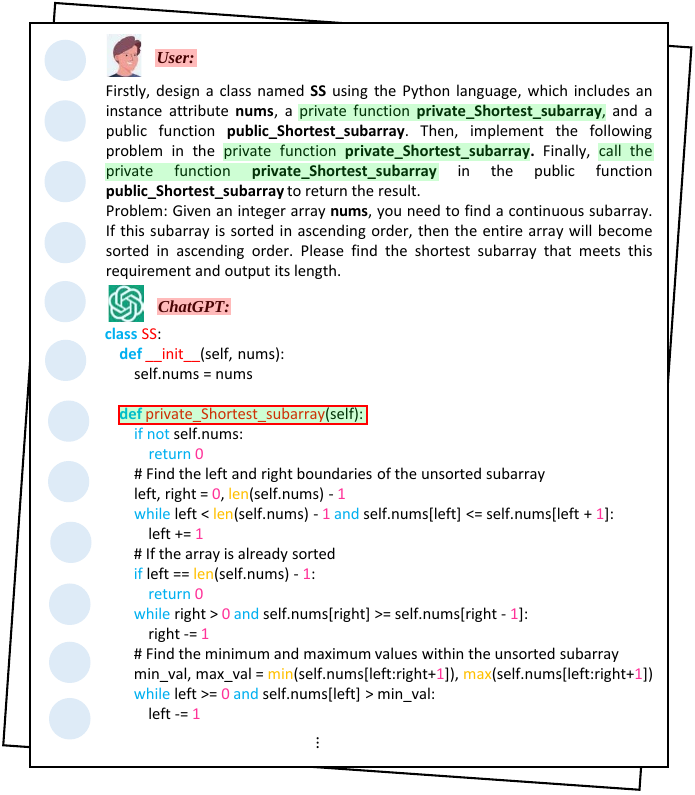}
    \caption{\textbf{The generation of private functions cannot be evaluated using \textit{pass@$k$}}. (We instructed ChatGPT~\cite{NEURIPS2022_b1efde53, openai2023gpt4} model to generate the class \textit{\textbf{class} SS}, public function \textit{public\_Shortest\_subarray}, and private function \underline{\textit{def \_\_private\_Shortest\_subarray}} based on a given prompt and implement the corresponding requirements within the functions. However, ChatGPT does not generate the private functions named \underline{\textit{private\_Shortest\_subarray}} outlined in the \sethlcolor{red}\hl{red box}.)}
    \label{fig:passk_shortcoming}
\end{figure}
\section{Evaluation Framework}
\subsection{Overview}
Existing code generation benchmarks in Table~\ref{tab:existing_benchmarks} for are confined to FP and do not involve essential concepts and features of OOP. We take the frequently used benchmarks, HumanEval~\cite{chen2021evaluating} and MBPP~\cite{austin2021program} in Table~\ref{tab:existing_benchmarks}, as examples. They primarily evaluate the capabilities of LLMs in FP. The detailed descriptions of HumanEval and MBPP are provided in Appendix~\ref{sec:appendix_motivate}. If we use existing benchmarks in Table~\ref{tab:existing_benchmarks} for evaluation, it does not show the capability of LLMs in OOP, as illustrated in Figure~\ref{fig:compcare_motivate}, that is, the seemingly decent LLMs (on FP tasks) perform relatively worse on OOP tasks. In addition, existing code generation evaluation metrics primarily use \textit{pass@$k$} to evaluate the executability of the generated code. However, using the \textit{pass@$k$} metric can not reflect whether LLMs generate concepts and features related to OOP, as illustrated in Figure~\ref{fig:passk_shortcoming}. Therefore, \textit{pass@$k$} can not objectively and fairly reflect the OOP capabilities of LLMs.

\begin{figure*}[!t]
    \centering
    \includegraphics[scale=0.65]{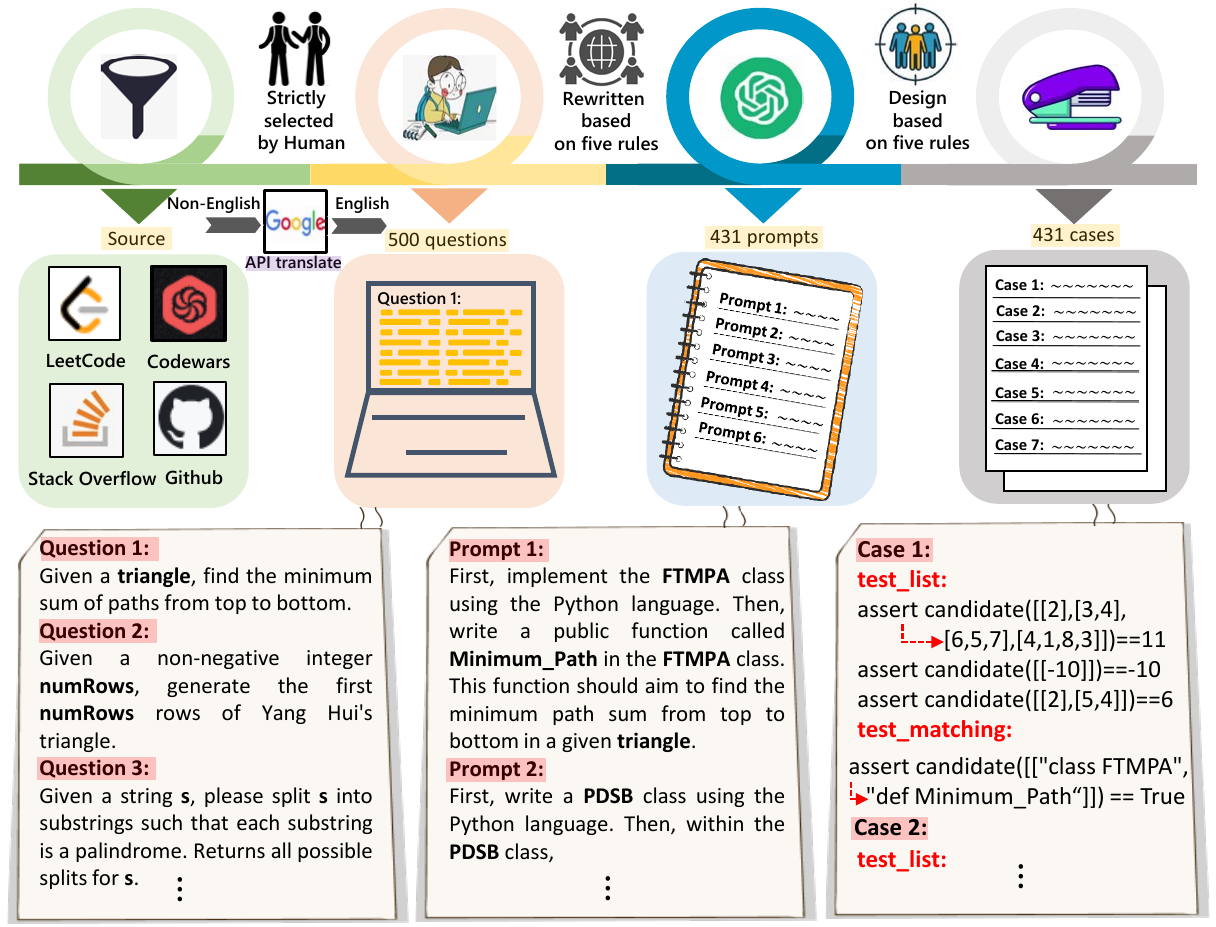}
    \caption{\textbf{The construction process of our object-oriented programming (OOP) benchmark}.}
    \label{fig:oop_construction}
\end{figure*}

As a result, we established an OOP benchmark and proposed the evaluation metric \textit{pass@$k$} for OOP. The process for constructing the OOP benchmark is illustrated in Figure~\ref{fig:oop_construction}.

\subsection{Building OOP Benchmarks}
\label{sec:building_oop}
\paragraph{Data Filtering.} The training data for current LLMs mostly comes from the internet. If we directly evaluate LLMs using existing OOP data from the web, it would not reflect the OOP capabilities of LLMs. Therefore, we first \textbf{rigorously selected $500$ natural language description-based problems or requirements based on Python} from platforms like LeetCode~\footnote{\url{https://leetcode.com/}}, open-source repositories on GitHub~\footnote{\url{https://github.com/}}, Stack Overflow~\footnote{\url{https://stackoverflow.com/}}, and Codewars~\footnote{\url{https://www.codewars.com/}}. These $500$ questions or requirements only are limited to FP and do not involve concepts and features related to OOP. 

\paragraph{Human Rewritten.} Subsequently, we manually rewrite the collected 500 questions or requirements by adhering to the following rules:
\begin{enumerate}
    \item Designing, based on the problems or requirements, with relevant OOP concepts and features, e.g., class names, inheritance name (i.e., parent class name), encapsulation methods name (i.e., public function name and private function names), etc.
    \item Related problems or requirements are implemented within the public function and private function of the class while ensuring the encapsulation of that implementation.
    \item Convert the variables associated with problems or requirements into class attribute variables, ensuring that these variables are accessible in both public and private functions.
    \item If the implementation of problems or requirements is placed within the private function of the class, it is necessary to design a corresponding public function for access.
    \item The rewritten OOP relevant problems or requirements can be successfully implemented and accessed through objects.
\end{enumerate}

Following the five rules mentioned above, we conducted a standardized rewriting of the $500$ Python-based problems or requirements. 

\paragraph{Case Design.} Finally, we designed corresponding test cases to evaluate OOP. 
Finally, we \textbf{obtained 431 samples of OOP}, as shown in Figure~\ref{fig:oop_construction}. The specific construction details of OOP are provided in Appendix~\ref{sec:appendix_level_classification}.

\paragraph{Level Classification.} Given the difficulty nature of programming, we divided the designed OOP benchmark into three levels: \textbf{Simple-level OOP}, \textbf{Moderate-level OOP}, and \textbf{Difficult-level OOP}, as shown in Figure~\ref{fig:oop_class_example}. 

Simple-level OOP has $77$ program samples, and includes only class, and public function. Moderate-level OOP builds upon simple-level OOP by adding attribute variables and private functions, and has $179$ program samples. Nevertheless,
the difficult-level OOP is based on the Simple-level of OOP, and adds inheritance, polymorphism and other related concepts and features of OOP. There are a total of $175$ program samples for difficult-level OOP. Although private functions are not involved in the difficulty level, the problems or requirements in difficult-level OOP are more complex and varied. Using such a level of classification, we can not only evaluate the performance of existing LLMs in OOP but also analyze the shortcomings of LLMs, which allows us to better unearth the potential of LLMs in OOP. Using this approach makes it more convenient for us to improve the OOP performance of LLMs.



\subsection{Evaluation Metrics \textit{Pass@$o$}}
To evaluate whether LLMs generate concepts and features related to OOP, i.e., generated subclass name, parent class name, private function name and public function name, etc, in the programming language, we proposed a \textit{pass@$o$} metric based on OOP. The \textit{pass@$o$} metric adds keyword points matching between natural language with programming language based on the \textit{pass@$k$}, i.e.,
\begin{align}
\label{eq:total_add_pass_matching}
&\quad\quad\quad\quad\,\,\,\,\alpha = \sum_{i=1}^{n} f\left(X_i\right),\nonumber\\
&where f(X_i)= \\
&\left\{
\begin{aligned}
    1,& \, if \, utf\left(X_i\right) \, passed \, and \, \sum_{j}^{m} x_j \exists X_i \\
    0,& \, \mathrm{otherwise}
\end{aligned}
\right., \nonumber
\end{align}
\begin{align}
\label{eq:mertric_pass@o}
\textit{pass@$o$}&:=\mathop{\mathbb{E}}_{Problems} \left[1-\frac{{\binom{n-\alpha}{o}}}{\binom{n}{o}} \right],
\end{align}
In Eq. (\ref{eq:total_add_pass_matching}), $n$ represents the number of code generations for a given problem; $X_i$ represents the $i$-${th}$ generated program code; $\alpha$ represents the quantity of $n$ generated codes passing tests and matches; $ut\left(\cdot\right)$ denotes the unit test function; $m$ represents the number of keyword points in the current $prompt$; and $x_j$ represents the $j$-${th}$ keyword points in natural language. In Eq. (\ref{eq:mertric_pass@o}), $o\leq n$.

The \textit{pass@$o$} metric not only optimizes the limitations of \textit{pass@$k$} evaluation but also objectively and fairly reflects the OOP performance of LLMs.

\section{Experiments}
\subsection{Experimental Setup}
\paragraph{Evaluated LLMs} In the OOP task, we conduct experiments on $23$ mainstream LLMs. These models include both general LLMs, i.g., ChatGPT~\cite{NEURIPS2022_b1efde53, openai2023gpt4}, Llama2~\cite{touvron2023llama}, InternLm~\cite{team2023internlm}, MPT~\cite{MosaicML2023Introducing}, DeepSeek~\cite{deepseekai2024deepseek}, Falcon~\cite{almazrouei2023falcon}, Qwen~\cite{bai2023qwen}, Yi~\footnote{\url{https://01.ai/cn}} and code-specialized LLMs, e.g., CodeLlama~\cite{roziere2023code}, WizardCoder~\cite{luo2023wizardcoder},
StarCoder~\cite{li2023starcoder}, as shown in Table~\ref{tab:models_overview}. The details description of $24$ LLMs are shown in Appendix~\ref{sec:appendix_LLMs}.


\paragraph{Parameter Settings.} In the experiment, we followed the settings on Llama2~\cite{touvron2023llama}, configuring the temperature to $0.1$ and $0.8$ for code generation. The remaining parameters $(top-p=0.95, n=200, o\le n)$, consistently remained unchanged. We evaluate the OOP benchmark on eight NVIDIA A100 GPUs using the vllm~\cite{kwon2023efficient} 0.2.1.post1 framework~\footnote{\url{https://github.com/vllm-project/vllm}}.

\paragraph{Metrics.} In terms of evaluation metrics, we use for \textit{pass@$k$} and the proposed \textit{pass@$o$} metrics. 
\begin{table*}[!t]
  \centering
    \resizebox{1.\linewidth}{!}{
    \begin{tabular}{clccccccccc}
    \toprule
    \multicolumn{2}{c}{\multirow{2}[2]{*}{\textbf{Model}}} & \multicolumn{3}{c}{\textbf{1}} & \multicolumn{3}{c}{\textbf{80}} & \multicolumn{3}{c}{\textbf{100}} \\
    \multicolumn{2}{c}{} & \textit{pass@$k$} & \textit{pass@$o$} & $\boldsymbol{\Delta}\left(\downarrow\right)$ & \textit{pass@$k$} & \textit{pass@$o$} & $\boldsymbol{\Delta}\left(\downarrow\right)$ & \textit{pass@$k$} & \textit{pass@$o$} & $\boldsymbol{\Delta}\left(\downarrow\right)$\\
    \midrule
    \multirow{15}[1]{*}{General}
    & Falcon-7b & $0.01$  & $0.00$ & $ \textcolor{red}{\textbf{-0.01}}$  & $0.37$  & $0.19$ & ${\textcolor{red}{\textbf{-0.18}}}$ & $0.47$  & $0.23$ & $\textcolor{red}{\textbf{-0.24}}$ \\
    & Falcon-40b & $0.01$  & $0.00$ & $\textcolor{red}{\textbf{-0.01}}$ & $2.90$  & $1.11$ & $\textcolor{red}{\textbf{-1.79}}$  & $3.42$  & $1.26$ & $\textcolor{red}{\textbf{-2.16}}$\\
    & Llama2-7b & $0.01$  & $0.01$ 
    & $\textcolor{red}{\textbf{-0.00}}$  & $4.02$  & $1.72$ & $\textcolor{red}{\textbf{-2.30}}$  & $4.62$  & $1.94$ & $\textcolor{red}{\textbf{-2.68}}$ \\
    & InternLm-7b & $0.03$  & $0.02$ & $\textcolor{red}{\textbf{-0.01}}$ & $1.04$  & $0.52$ & $\textcolor{red}{\textbf{-0.52}}$  & $1.22$  & $0.58$ & $\textcolor{red}{\textbf{-0.64}}$ \\
    & Yi-6b & $0.07$  & $0.01$ & $\textcolor{red}{\textbf{-0.06}}$ & $5.07$  & $1.67$ & $\textcolor{red}{\textbf{-3.40}}$ & $6.00$  & $1.98$ & $\textcolor{red}{\textbf{-4.02}}$ \\
    & Llama2-13b & $0.09$  & $0.06$ & $\textcolor{red}{\textbf{-0.03}}$  & $7.28$  & $2.17$ & $\textcolor{red}{\textbf{-5.11}}$  & $8.24$  & $2.41$ & $\textcolor{red}{\textbf{-5.83}}$ \\
    & MPT-7b & $0.28$  & $0.02$ & $\textcolor{red}{\textbf{-0.26}}$ & $4.77$  & $1.27$ & $\textcolor{red}{\textbf{-3.50}}$  & $5.50$  & $1.46$ & $\textcolor{red}{\textbf{-4.04}}$ \\
    & Qwen-7b & $0.94$  & $0.61$ & $\textcolor{red}{\textbf{-0.33}}$  & $15.02$  & $5.68$ & $\textcolor{red}{\textbf{-9.34}}$ & $16.35$  & $5.83$ & $\textcolor{red}{\textbf{-10.52}}$ \\
    & Qwen-14b & $1.52$  & $0.75$ & $\textcolor{red}{\textbf{-0.77}}$  & $26.28$  & $10.58$ & $\textcolor{red}{\textbf{-15.70}}$  & $28.10$  & $11.48$ & $\textcolor{red}{\textbf{-16.62}}$ \\
    & DeepSeek-7b & $1.53$  & $0.50$ & $\textcolor{red}{\textbf{-1.03}}$  & $16.83$  & $7.72$ & $\textcolor{red}{\textbf{-9.11}}$ & $18.70$  & $8.70$ & $\textcolor{red}{\textbf{-10.00}}$ \\
    & Yi-34b & $2.20$  & $1.09$ & $\textcolor{red}{\textbf{-1.11}}$  & $21.96$  & $8.43$ & $\textcolor{red}{\textbf{-13.53}}$ & $23.68$  & $9.22$ & $\textcolor{red}{\textbf{-14.46}}$ \\
    & \cellcolor[rgb]{.906,  .902,  .902}Llama2-70b & \cellcolor[rgb]{.906,  .902,  .902}$3.55$  & \cellcolor[rgb]{.906,  .902,  .902}$1.25$ & \cellcolor[rgb]{.906,  .902,  .902}$\textcolor{red}{\textbf{-2.30}}$ & \cellcolor[rgb]{.906,  .902,  .902}$21.01$  & \cellcolor[rgb]{.906,  .902,  .902}$9.97$ & \cellcolor[rgb]{.906,  .902,  .902}$\textcolor{red}{\textbf{-11.04}}$ & \cellcolor[rgb]{.906,  .902,  .902}$23.14$  & \cellcolor[rgb]{.906,  .902,  .902}$11.16$ & \cellcolor[rgb]{.906,  .902,  .902}$\textcolor{red}{\textbf{-11.98}}$ \\
    & \cellcolor[rgb]{.906,  .902,  .902}DeepSeek-67b & \cellcolor[rgb]{.906,  .902,  .902}$8.02$  & \cellcolor[rgb]{.906,  .902,  .902}$3.71$ & \cellcolor[rgb]{.906,  .902,  .902}$\textcolor{red}{\textbf{-3.95}}$  & \cellcolor[rgb]{.906,  .902,  .902}$49.31$  & \cellcolor[rgb]{.906,  .902,  .902}$27.42$ & \cellcolor[rgb]{.906,  .902,  .902}$\textcolor{red}{\underline{\textbf{-21.89}}}$ & \cellcolor[rgb]{.906,  .902,  .902}$51.60$  & \cellcolor[rgb]{.906,  .902,  .902}$29.47$ & \cellcolor[rgb]{.906,  .902,  .902}$\textcolor{red}{\underline{\textbf{-22.13}}}$ \\
     &\cellcolor[rgb]{.906,  .902,  .902} Qwen-72b &\cellcolor[rgb]{.906,  .902,  .902} $11.20$  &\cellcolor[rgb]{.906,  .902,  .902} $4.62$ & \cellcolor[rgb]{.906,  .902,  .902}$\textcolor{red}{\textbf{-6.58}}$ & \cellcolor[rgb]{.906,  .902,  .902}$57.48$  & \cellcolor[rgb]{.906,  .902,  .902}$35.70$ & \cellcolor[rgb]{.906,  .902,  .902}$\textcolor{red}{\textbf{-21.78}}$ & \cellcolor[rgb]{.906,  .902,  .902}$59.52$  & \cellcolor[rgb]{.906,  .902,  .902}$37.83$ & \cellcolor[rgb]{.906,  .902,  .902}$\textcolor{red}{\textbf{-21.69}}$ \\
    & ChatGPT & $\textbf{42.88}$  & ${\textbf{15.69}}$ & $\textcolor{red}{\underline{\textbf{-27.19}}}$  & $\textbf{75.71}$ & ${\textbf{58.28}}$ & $\textcolor{red}{\textbf{-17.43}}$ & $\textbf{76.20}$ & ${\textbf{59.80}}$ & $\textcolor{red}{\textbf{-16.40}}$\\
    \hdashline
    \multirow{7}[1]{*}{Specialized} 
    & GPT\_BigCode & $0.10$  & $0.06$ & $\textcolor{red}{\textbf{-0.04}}$  & $7.00$  & $2.58$ & $\textcolor{red}{\textbf{-4.42}}$  & $8.01$  & $2.92$ & $\textcolor{red}{\textbf{-5.09}}$ \\
    & CodeLlama-7b & $2.67$  & $1.20$ & $\textcolor{red}{\textbf{-1.47}}$ & $24.09$  & $9.16$ & $\textcolor{red}{\textbf{-14.93}}$ & $25.72$  & $9.92$ & $\textcolor{red}{\textbf{-15.80}}$ \\
    & CodeLlama-13b-Python & $2.80$  & $1.03$ & $\textcolor{red}{\textbf{-1.77}}$ & $36.34$  & $17.22$ & $\textcolor{red}{\textbf{-19.12}}$  & $38.75$  & $18.96$ & $\textcolor{red}{\textbf{-19.79}}$ \\
    & StarCoder & $4.61$  & $1.26$ & $\textcolor{red}{\textbf{-3.35}}$ & $28.67$  & $10.05$ & $\textcolor{red}{\textbf{-18.62}}$  & $30.44$  & $10.88$ & $\textcolor{red}{\textbf{-19.56}}$ \\
    & CodeLlama-7b-Python & $4.68$  & $1.27$ & $\textcolor{red}{\textbf{-3.41}}$ & $28.68$  & $12.90$ & $\textcolor{red}{\textbf{-15.78}}$ & $30.33$  & $14.07$ & $\textcolor{red}{\textbf{-16.26}}$ \\
    & \cellcolor[rgb]{.906,  .902,  .902}CodeLlama-34b & \cellcolor[rgb]{.906,  .902,  .902}$6.24$  & \cellcolor[rgb]{.906,  .902,  .902}$1.58$ & \cellcolor[rgb]{.906,  .902,  .902}$\textcolor{red}{\underline{\textbf{-4.66}}}$ & \cellcolor[rgb]{.906,  .902,  .902}$\textbf{46.31}$  & \cellcolor[rgb]{.906,  .902,  .902}$\textbf{22.59}$ & \cellcolor[rgb]{.906,  .902,  .902}$\textcolor{red}{\underline{\textbf{-23.72}}}$ & \cellcolor[rgb]{.906,  .902,  .902}$\textbf{49.01}$  & \cellcolor[rgb]{.906,  .902,  .902}$\textbf{24.68}$ & \cellcolor[rgb]{.906,  .902,  .902}$\textcolor{red}{\underline{\textbf{-24.33}}}$ \\
    & WizardCoder-15b & $6.83$  & $\textbf{3.02}$ & $\textcolor{red}{\textbf{-3.81}}$ & $28.10$  & $9.50$ & $\textcolor{red}{\textbf{-18.60}}$ & $29.41$  & $10.01$ & $\textcolor{red}{\textbf{-19.40}}$\\
    & CodeLlama-13b & $\textbf{6.87}$  & $2.92$ & $\textcolor{red}{\textbf{-3.95}}$ & $32.69$  & $12.20$ & $\textcolor{red}{\textbf{-20.49}}$  & $34.53$  & $13.11$ & $\textcolor{red}{\textbf{-21.42}}$ \\
    \bottomrule
    \end{tabular}%
    }
    \caption{\textbf{Performance of $23$ large language models (LLMs) on object-oriented programming (OOP) tasks}. We also reported the differences in evaluation results between \textit{pass@k} and \textit{pass@o}. (All LLMs are evaluated in zero-shot fashion. For \textit{pass@$100$} and \textit{pass@$80$} scores, we use a temperature of $0.8$ and top-$p$=$0.95$. For \textit{pass@$1$} scores, we use a temperature of $0.1$ and top-$p$=$0.95$. The best results are highlighted in black bold; \sethlcolor{red}\hl{Red} indicates the differences evaluated using the \textit{pass@o} and \textit{pass@k} metrics; \underline{Underlined} indicates the maximum disparities evaluated between \textit{pass@o} and \textit{pass@k} metrics; \sethlcolor{lightgray}\hl{Gray} indicates models with a larger number of parameters.)}
  \label{tab:all_OOP_scores}%
\end{table*}%

\subsection{Overall Evaluation Result}
The evaluation results of the LLMs with temperatures set to 0.1 and 0.8 are presented in Table~\ref{tab:all_OOP_scores}, respectively. 
From the experimental results, We have obtained the following conclusions: 

\paragraph{The OOP capabilities of the existing LLMs fall far short of the ideal state} In Table~\ref{tab:all_OOP_scores}, we can observe that LLMs with strong coding capabilities (e.g., WizardCoder-15b, CodeLlama-7b-Python, and CodeLlama-13b, achieved scores of $58.12$, $40.48$, and $35.07$, respectively, in the HumanEval code leaderboard~\footnote{\url{https://huggingface.co/spaces/bigcode/bigcode-models-leaderboard}}), exhibit performance in OOP benchmarks that falls significantly short of the ideal state. WizardCoder-15b, CodeLlama-7b-Python, and CodeLlama-13b scored $3.02$, $1.27$, and $2.92$, respectively, on the OOP benchmark at \textit{pass@$1$}. Their scores on \textit{pass@$100$} were also $10.01$, $14.07$, and $13.11$, respectively. Even the current ChatGPT model with strong general capabilities scores $15.69$ on \textit{pass@$1$} and $59.80$ on \textit{pass@$100$}. The results indicate that the untapped potential of existing LLMs in OOP has not been fully explored.

\paragraph{Limitations of \textit{pass@$k$} evaluated OOP} The scores from Table~\ref{tab:all_OOP_scores} indicate that using \textit{pass@$k$} does not objectively reflect the OOP performance of LLMs, e.g., the WizardCoder-15b model achieves scores of $6.83$, $28.10$, and $29.41$ using \textit{pass@$k$}, while its scores drop to $3.02$, $9.50$, and $10.01$ when using \textit{pass@$o$}. The evaluation scores of other LLMs using \textit{pass@$o$} in Table~\ref{tab:all_OOP_scores} showed a decline, once again proving the limitations of \textit{pass@$k$} in evaluation OOP.

\begin{figure*}[!t]
    \centering
    \includegraphics[width=\textwidth]{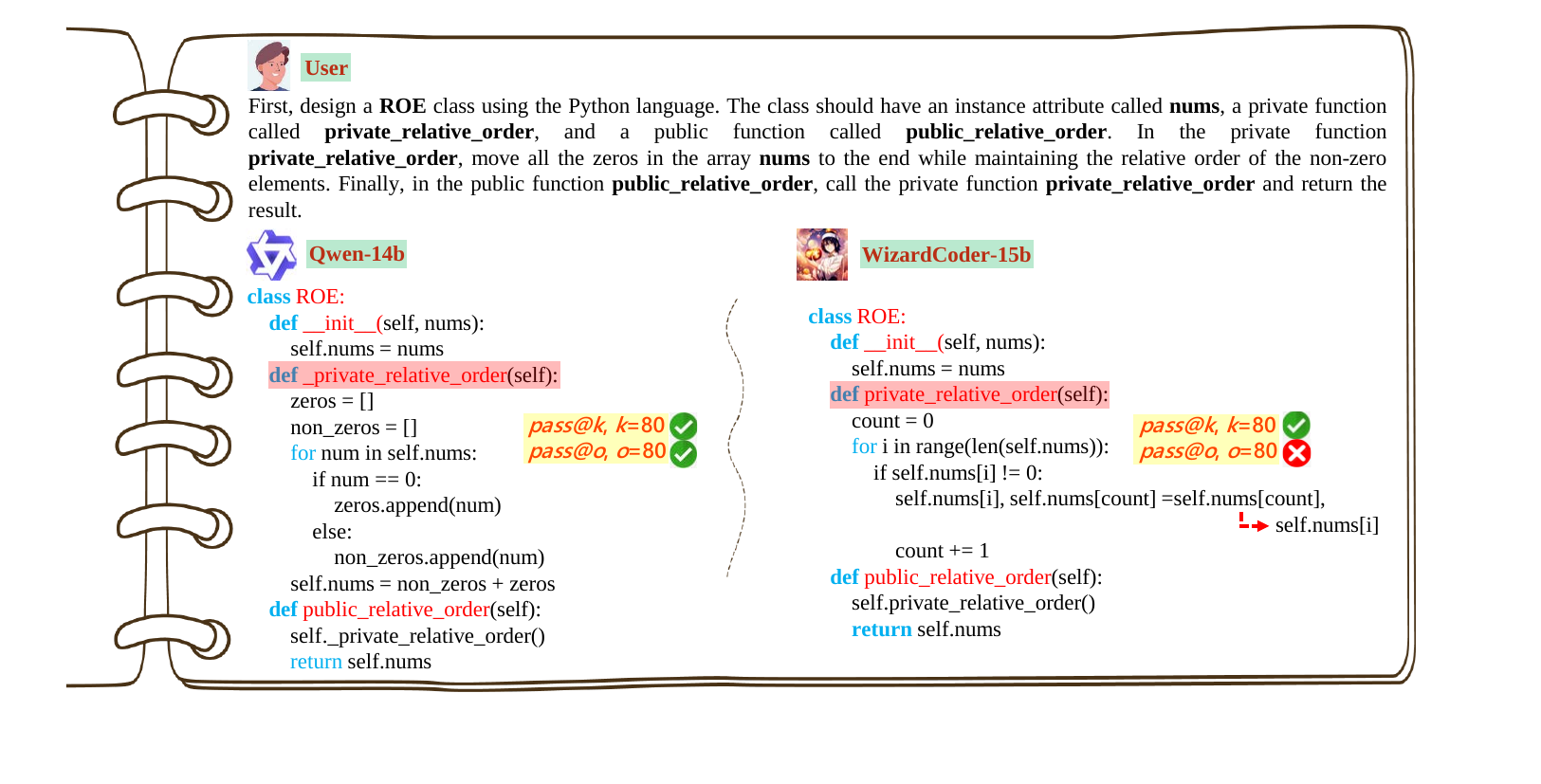}
    \caption{\textbf{The case comparison of generation results between Qwen-14b and WizardCoder-15b in the OOP benchmark}. We see: 1) Qwen-14b can accurately generate private functions, while WizardCoder-15b cannot accurately generate private functions; 2) The results generated by Qwen-14b and WizardCoder-15b can both pass the evaluation using \textit{pass@$k$}; 3) The results generated by Qwen-14b can pass the evaluation using \textit{pass@$o$}, but the results generated by WizardCoder-15b cannot pass the evaluation using \textit{pass@$o$}.}
    \label{fig:passk_passo}
\end{figure*}

In addition, we also observed a significant phenomenon, e.g., when evaluated using \textit{pass@$k$}, Qwen-14b (score $26.28$) scored lower than WizardCoder-15b (score $28.10$) on \textit{pass@$80$}. However, when evaluated using \textit{pass@$o$}, Qwen-14b (score 
 $10.58$) scored higher than WizardCoder-15b (score $9.50$) on \textit{pass@$80$}. Analyzing the experimental results of Qwen-14b and WizardCoder-15b, we observed that when evaluated using \textit{pass@$o$}, Qwen-14b outperforms WizardCoder-15b in terms of the ability to correctly generate OOP concepts and feature keywords, as illustrated in Figure~\ref{fig:passk_passo}. It also reiterates that \textit{pass@$k$} cannot objectively and fairly reflect the evaluation results of OOP.

\paragraph{A larger model scale does not necessarily perform better on \textit{pass@$1$}}
In Table~\ref{tab:all_OOP_scores}, when evaluated using \textit{pass@$o$}, CodeLlama-34b scores $1.58$ on \textit{pass@$1$}, whereas CodeLlama-13b scores $2.92$ on \textit{pass@$1$}. Additionally, CodeLlama-13b-Python scores $1.03$ on \textit{pass@$1$}, while the corresponding CodeLlama-7b-Python scores $1.27$ on \textit{pass@$1$}. 
However, CodeLlama-7b scores $1.20$ on \textit{pass@$1$}, which is lower than the score achieved by CodeLlama-13b.
The scores of CodeLlama-7b, CodeLlama-13b, CodeLlama-34b, CodeLlama-7b-Python, and CodeLlama-13b-Python on \textit{pass@$1$} indicate that a larger model scale does not necessarily result in the highest scores on \textit{pass@$1$}.



\subsection{Different-level Evaluation Results}
Following the classification of OOP benchmarks in Section~\ref{sec:building_oop}, we conducted evaluations for three levels of OOP benchmarks, and the results are presented in Tables~\ref{tab:simple_level_OOP_scores},~\ref{tab:moderate_level_OOP_scores}, and~\ref{tab:difficult_level_OOP_scores}, respectively. we have drawn the following conclusions:

\paragraph{LLMs perform better at the simple-level OOP compared to the moderate-level and difficult-level OOP} From the simple-level OOP evaluation results in Table~\ref{tab:simple_level_OOP_scores}, we can see that the evaluation results using \textit{pass@$k$} and \textit{pass@$o$} are the same. It also indicates that LLMs can comprehend the fundamental concepts and features of OOP, e.g., class, and encapsulation methods (i.e., public function). However, in Tables~\ref{tab:moderate_level_OOP_scores} and~\ref{tab:difficult_level_OOP_scores}, LLMs exhibit a weaker understanding of concepts and features related to OOP, e.g., encapsulation methods (i.e., private function), inheritance, and polymorphism, and are unable to generate corresponding code accurately. Detailed descriptions are in Appendix~\ref{sec:appendix_Analysis_results}. 


\paragraph{ChatGPT has large gap in moderate-level usage using \textit{pass@$k$} and \textit{pass@$o$}}
From the results in Table~\ref{tab:moderate_level_OOP_scores}, 
We observe that with \textit{pass@$k$} evaluation, ChatGPT scores are $51.71$, $83.30$, and $83.67$, but with \textit{pass@$o$} evaluation, ChatGPT only achieves scores of $2.53$, $51.54$, and $54.78$.
We analyzed the moderate-level OOP results for ChatGPT, and found that its understanding of private functions is relatively poor.
When evaluated using \textit{pass@$k$}, a total of $5551$ codes can pass the test cases correctly. However, when evaluated using \textit{pass@$o$}, only $272$ codes can successfully pass the test cases. Among them, $5279$ codes fail to match the \textit{pass@$o$} criteria. Upon careful examination, we found that all these $5279$ codes resulted from errors generated by private functions
To further validate the authenticity of the experimental results, we randomly selected prompts corresponding to three error results. Subsequently, we input prompts of the erroneous results into the web version of ChatGPT for code generation, as illustrated in Figure~\ref{fig:Web_ChatGPT_code01}, ~\ref{fig:Web_ChatGPT_code02} and~\ref{fig:Web_ChatGPT_code03}. 
We found that the code generated by online ChatGPT~\footnote{\url{https://chat.openai.com/}} is also private function error. 

\paragraph{ChatGPT outperforms moderate-level in difficult-level evaluation results} 
According to the evaluation results from Tables~\ref{tab:simple_level_OOP_scores} and~\ref{tab:moderate_level_OOP_scores}, we observe that the performance of ChatGPT at the difficult level is stronger than at the moderate level. At the difficult-level OOP, ChatGPT scores are $19.70$, $71.83$, and $73.37$, whereas at the moderate-level OOP, ChatGPT scores are only $2.53$, $51.54$, and $54.78$.




\begin{figure}[!t]
    \centering
    \includegraphics[scale=0.45]{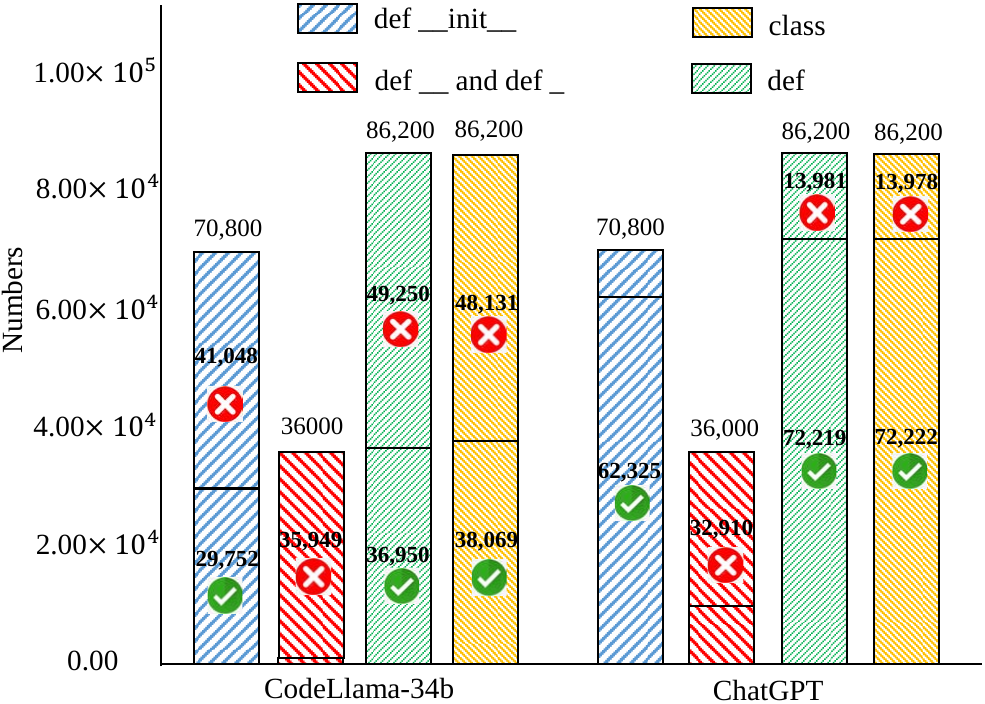}
    \caption{\textbf{Distribution of search results for ChatGPT and CodeLlama-34b.} (In program, ``\textbf{class}'' serves as the indicator for program \underline{\textit{class names}}. If the program does not contain a ``\textbf{class}'', it signifies an error in the generation of \underline{\textit{class names}} by the LLM. Similarly, it can be deduced that ``\textbf{def \_}'' and ``\textbf{def \_\_}'' serve as indicators for \underline{\textit{private function names}}; ``\textbf{def}'' signifies a \underline{\textit{public function name}}; and ``\textbf{def \_\_init\_\_}'' represents the indicator for \underline{\textit{attribute variables name}}. Moreover, In our OOP benchmark, the LLM should ideally generate at least 86,200 ``\textbf{class}'', 36,000 ``\textbf{def \_\_}'' or ``\textbf{def \_}'', 86,200 ``\textbf{def}'', and 70,800 ``\textbf{def \_\_init\_\_}''.)}
    \label{fig:CodeLlama34b_ChatGPT_error_type}
\end{figure}


\section{Discussion}
In this section, we will explore the reasons behind the generally lower scores of LLMs in OOP, as well as the applicability of the Chain-of-Thoughts (CoT) method to OOP.

\paragraph{Why LLMs score lower in OOP benchmarks?}
We use the experimental results of ChatGPT and CodeLlama-34b on \textit{pass@$1$} as examples for analysis. As we instruct LLMs to generate relevant class names, private function names, public function names, etc., We conducted searches using simple keywords, e.g., ``\textbf{class}'', ``\textbf{def \_}'', ``\textbf{def \_\_}'', ``\textbf{def \_\_init\_\_}'', and ``\textbf{def}'' on both CodeLlama-34b and ChatGPT results. A detailed description of the retrieval process is provided in Appendix~\ref{sec:retrieval_process}. We compiled and analyzed the distribution of retrieval ``\textbf{class}'', ``\textbf{def \_}'', ``\textbf{def \_\_}'', ``\textbf{def \_\_init\_\_}'', and ``\textbf{def}'', as shown in Figure~\ref{fig:CodeLlama34b_ChatGPT_error_type}, concluding that: 1) Weak knowledge, e.g., class, encapsulation methods, etc, of OOP in LLMs; 2) LLMs particularly lack cognition of private functions; 3) There is a certain degree of gap between CodeLlama-34b and ChatGPT. Specific example is shown in Figure~\ref{fig:CodeLlama34b_vs_ChatGPT_01}.


\paragraph{The applicability of CoT in OOP.}
Taking CodeLlama-13b, StarCoder, and WizardCoder-15b as examples, we respectively incorporate the few-shot, zero-shot CoT, and few-shot CoT methods to validate whether CoT approaches demonstrate applicability in OOP, as shown in Table~\ref{tab:cot_methods}.
We observed a significant improvement in the scores of LLMs in OOP when using the few-shot approach, e.g., CodeLlama-13b achieved scores of $14.50$, $48.13$, and $49.85$ using the few-shot method, representing improvements of $396.58\%$, $294.51\%$, and $280.24\%$, respectively, compared to the zero-shot method. In Table~\ref{tab:cot_methods}, we also observe that CodeLlama-13b achieves scores of $1.33$, $13.31$, and $14.62$ in zero-shot CoT, but its score at \textit{pass@$1$} is lower at $2.92$ compared to zero-shot. Additionally, StarCoder scores $0.25$, $6.58$, and $7.07$ in zero-shot CoT, which are lower than StarCoder scores in zero-shot at $1.26$, $10.05$, and $10.88$, respectively. The scores of the CodeLlama-13b, StarCoder, and WizardCoder-15b models on few-shot CoT are also lower than their scores on few-shot. We analyzed the experimental results of zero-shot and zero-shot CoT and found that using the CoT method introduces an illusion to the model, preventing it from directly generating the corresponding code, as illustrated in Figure~\ref{fig:zero_shot_vs_zero_shot_cot}. Therefore, it is necessary to integrate the concepts and features of OOP to design appropriate CoT strategies in order to enhance the effectiveness of generating OOP by LLMs. Appendix~\ref{sec:few_cot} provides detailed prompts for few-shot, zero-shot CoT, and few-shot CoT.
\section{Conclusion}
In this paper, we propose the first OOP evaluation benchmark based on Python, consisting of $431$ Python programs, encompassing key concepts and features of OOP, e.g., class, encapsulation methods, etc. Simultaneously, we propose the evaluation metric \textit{pass@$o$} for the OOP benchmark. \textit{pass@$o$} improves upon the limitations of \textit{pass@$k$} by matching keyword points between natural language with program language. We evaluate $23$ mainstream LLMs using the proposed OOP benchmark and \textit{pass@$o$} metric. Experimental results show that the current OOP of LLMs is far from ideal, which also reveals that LLMs have room for further improvement. Furthermore, Existing LLMs have a certain gap with ChatGPT in OOP. Moreover, we also investigate that applying some of the current improvement strategies directly to the OOP benchmark does not show significant improvement. In the future, we need to further strengthen the OOP knowledge of LLMs, especially regarding private functions. At the same time, we also hope that more researchers can contribute to the advancement of research in OOP.

\section*{Limitations} Our OOP benchmark has several limitations: (1) Our proposed OOP benchmark is based on the Python programming language and does not cover other OOP languages. (2) Given the incorporation of crucial concepts like polymorphism and inheritance in the OOP benchmark, it does not specifically address challenges associated with more intricate scenarios, e.g., multiple inheritance and overloading. (3) While OOP languages hold a significant share, non-OOP languages, e.g., C and Go languages, also play irreplaceable roles. In future work, we plan to consider expanding the OOP benchmark to cover a broader spectrum. Additionally, we encourage researchers to explore the potential of LLMs through evaluations based on the OOP benchmark.

\section*{Ethics Statement}
We take ethical considerations very seriously. This paper focuses on establishing benchmarks for OOP to analyze the performance of existing LLMs. Our research reveals that existing LLMs fall far short of ideal performance in OOP. We conducted experiments on open and publicly available LLMs and accurately and objectively report the findings and conclusions of this paper. Therefore, we believe that this study does not raise ethical concerns.
\bibliography{acl2023}
\bibliographystyle{acl_natbib}

\clearpage
\appendix

\begin{figure}[!t]
    \centering
    \subfigure[HumanEval.]{\includegraphics[width=0.48\textwidth]{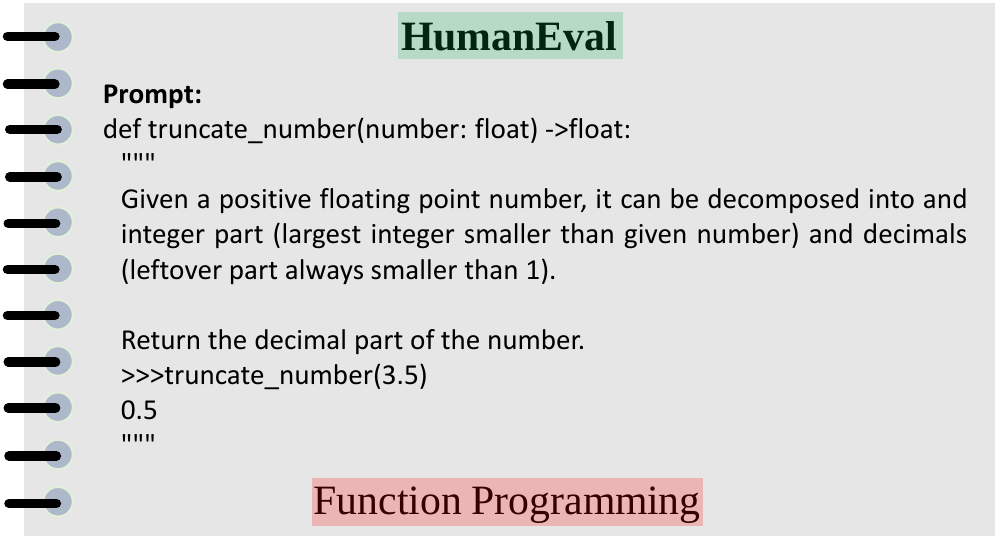}}
    \subfigure[MBPP.]{\includegraphics[width=0.48\textwidth]{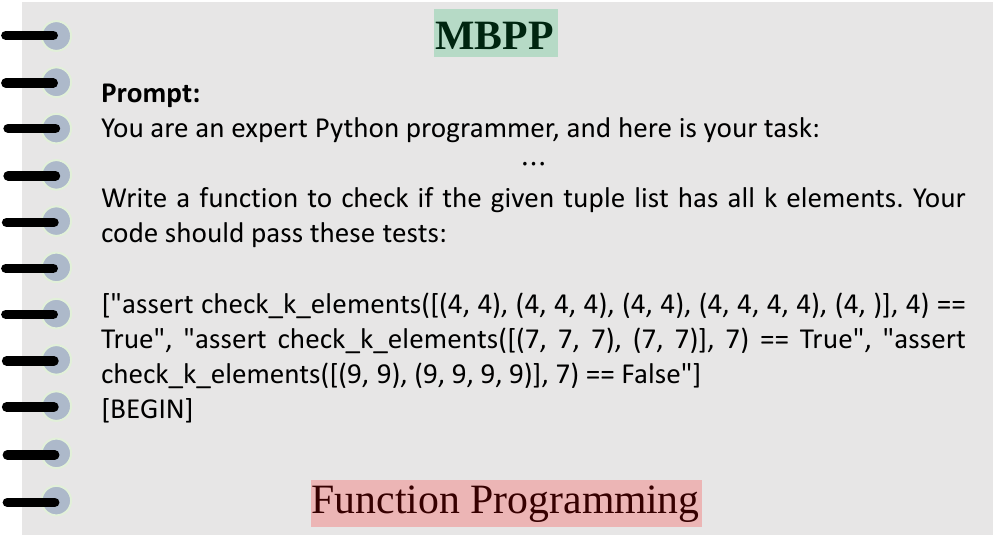}}
    \subfigure[OOP.]{\includegraphics[width=0.48\textwidth]{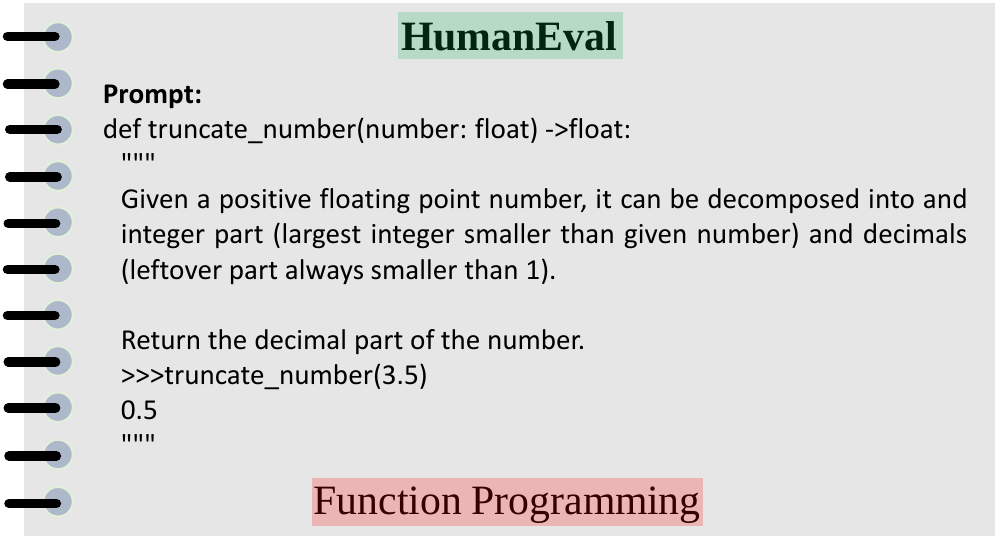}}
    \caption{\textbf{Differences between OOP benchmarks and HumanEval, as well as MBPP Benchmarks ($\dots$ indicates that the few-shot content in MBPP is omitted )}. We can see that: 1) the HumanEval benchmark requires models to complete based on the context within the function; 2) the MBPP benchmark directly requires models to generate based on prompt requirements; 3) However, our proposed OOP benchmark requirements are generated based on specified prompt as well as concepts and features of OOP. Therefore, HumanEval and MBPP do not reflect the concepts and features of OOP.}
    \label{fig:humaneval_and_mbpp_shortcoming}
\end{figure}


\begin{figure}[!t]
    \centering
    \subfigure[Simple-level.]{\includegraphics[width=0.48\textwidth]{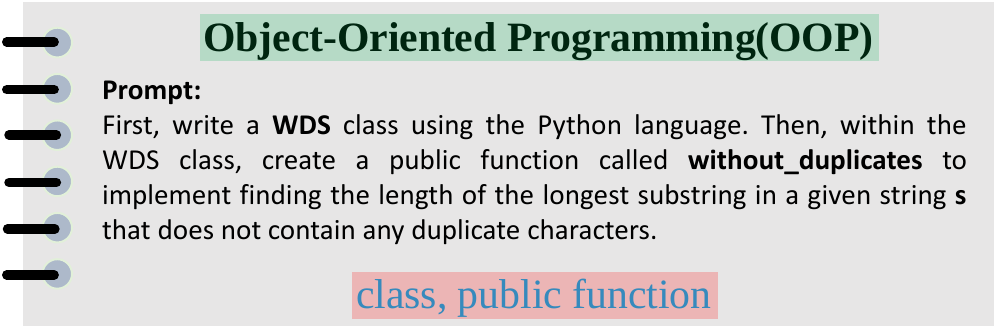}}
    \label{fig:simple_level}
    \hspace{0.02cm} 
    \subfigure[Moderate-level.]{\includegraphics[width=0.48\textwidth]{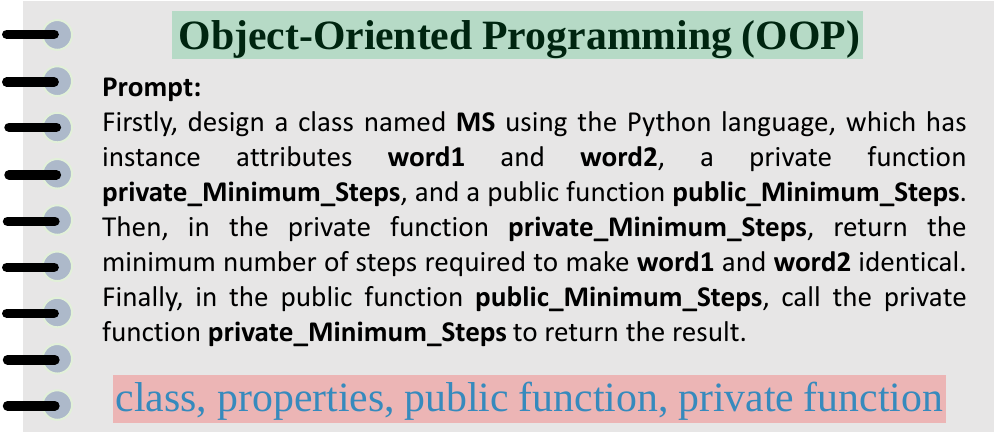}}
    \label{fig:moderate_level}
    \hspace{0.02cm} 
    
    \subfigure[Difficult-level.]{\includegraphics[width=0.48\textwidth]{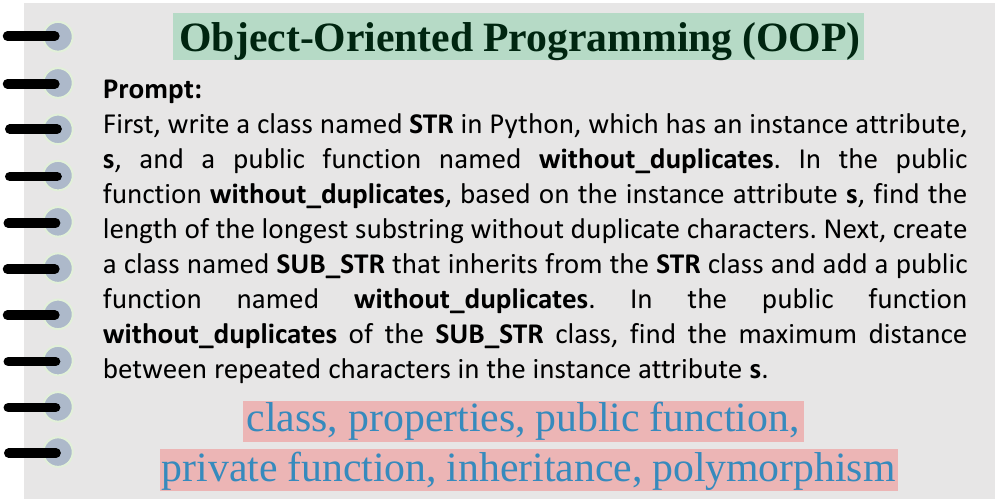}}
    \label{fig:difficult_level_02}
    \caption{\textbf{Examples of different levels for object-oriented programming (OOP) tasks}.}
    \label{fig:oop_class_example}
\end{figure}

\begin{table*}[!t]
  \centering
    \resizebox{1.\linewidth}{!}{
    \begin{tabular}{clccccccccc}
    \toprule
    \multicolumn{2}{c}{\multirow{2}[2]{*}{\textbf{Model}}} & \multicolumn{3}{c}{\textbf{1}} & \multicolumn{3}{c}{\textbf{80}} & \multicolumn{3}{c}{\textbf{100}} \\
    \multicolumn{2}{c}{} & \textit{pass@$k$} & \textit{pass@$o$} & $\boldsymbol{\Delta}\left(\downarrow\right)$ & \textit{pass@$k$} & \textit{pass@$o$} & $\boldsymbol{\Delta}\left(\downarrow\right)$ & \textit{pass@$k$} & \textit{pass@$o$} & $\boldsymbol{\Delta}\left(\downarrow\right)$\\
    \midrule
    \multirow{15}[1]{*}{General}
    & Falcon-7b & $0.00$  & $0.00$ & $\textcolor{red}{\textbf{-0.00}}$  & $1.04$  & $1.04$  & $\textcolor{red}{\textbf{-0.00}}$ & $1.30$ & $1.30$ & $\textcolor{red}{\textbf{-0.00}}$ \\
    & Falcon-40b & $0.00$  & $0.00$ & $\textcolor{red}{\textbf{-0.00}}$ & $5.10$  & $5.10$ & $\textcolor{red}{\textbf{-0.00}}$  & $5.68$ & $5.68$ & $\textcolor{red}{\textbf{-0.00}}$\\
    & Yi-6b & $0.00$  & $0.00$ & $\textcolor{red}{\textbf{-0.00}}$ & $5.87$  & $5.87$ & $\textcolor{red}{\textbf{-0.00}}$  & $6.76$ & $6.76$ & $\textcolor{red}{\textbf{-0.00}}$ \\
    & Llama2-7b & $0.03$  & $0.03$ & $\textcolor{red}{\textbf{-0.00}}$  & $9.56$  & $9.56$ & $\textcolor{red}{\textbf{-0.00}}$  & $10.77$ & $10.77$ & $\textcolor{red}{\textbf{-0.00}}$ \\
    & InternLm-7b & $0.09$  & $0.09$ & $\textcolor{red}{\textbf{-0.00}}$  & $2.87$  & $2.87$ & $\textcolor{red}{\textbf{-0.00}}$ & $3.21$  & $3.21$ & $\textcolor{red}{\textbf{-0.00}}$ \\
    & MPT-7b & $0.13$  & $0.13$ & $\textcolor{red}{\textbf{-0.00}}$ & $7.03$  & $7.03$ & $\textcolor{red}{\textbf{-0.00}}$ & $8.13$  & $8.13$ & $\textcolor{red}{\textbf{-0.00}}$ \\
    & Llama2-13b & $0.32$  & $0.32$ & $\textcolor{red}{\textbf{-0.00}}$ & $12.05$ & $12.05$ & $\textcolor{red}{\textbf{-0.00}}$ & $13.39$ & $13.39$ & $\textcolor{red}{\textbf{-0.00}}$ \\
    & DeepSeek-7b & $0.72$  & $0.72$ & $\textcolor{red}{\textbf{-0.00}}$ & $24.03$ & $24.03$ & $\textcolor{red}{\textbf{-0.00}}$ & $26.12$ & $26.12$ & $\textcolor{red}{\textbf{-0.00}}$ \\
    & Qwen-7b & $3.36$  & $3.36$ & $\textcolor{red}{\textbf{-0.00}}$ & $30.53$ & $30.53$ & $\textcolor{red}{\textbf{-0.00}}$ & $31.24$ & $31.24$ & $\textcolor{red}{\textbf{-0.00}}$ \\
    & Yi-34b & $3.41$  & $3.41$ & $\textcolor{red}{\textbf{-0.00}}$ & $26.16$ & $26.16$ & $\textcolor{red}{\textbf{-0.00}}$ & $27.63$ & $27.63$ & $\textcolor{red}{\textbf{-0.00}}$ \\
    & \cellcolor[rgb]{.906,  .902,  .902} Llama2-70b & \cellcolor[rgb]{.906,  .902,  .902}$3.79$  & \cellcolor[rgb]{.906,  .902,  .902}$3.79$ & $\cellcolor[rgb]{.906,  .902,  .902}\textcolor{red}{\textbf{-0.00}}$ & \cellcolor[rgb]{.906,  .902,  .902}$27.15$ & \cellcolor[rgb]{.906,  .902,  .902}$27.15$ & \cellcolor[rgb]{.906,  .902,  .902}$\textcolor{red}{\textbf{-0.00}}$ & \cellcolor[rgb]{.906,  .902,  .902}$29.52$ & \cellcolor[rgb]{.906,  .902,  .902}$29.52$ & \cellcolor[rgb]{.906,  .902,  .902}$\textcolor{red}{\textbf{-0.00}}$ \\
    & Qwen-14b & $4.06$  & $4.06$ & $\textcolor{red}{\textbf{-0.00}}$ & $36.89$ & $36.89$ & $\textcolor{red}{\textbf{-0.00}}$ & $37.87$ & $37.87$ & $\textcolor{red}{\textbf{-0.00}}$ \\
    & \cellcolor[rgb]{.906,  .902,  .902}DeepSeek-67b & \cellcolor[rgb]{.906,  .902,  .902}$10.36$  & \cellcolor[rgb]{.906,  .902,  .902}$10.36$ & $\cellcolor[rgb]{.906,  .902,  .902}\textcolor{red}{\textbf{-0.00}}$  & \cellcolor[rgb]{.906,  .902,  .902}$52.75$ & \cellcolor[rgb]{.906,  .902,  .902}$52.75$ & \cellcolor[rgb]{.906,  .902,  .902}$\textcolor{red}{\textbf{-0.00}}$ & \cellcolor[rgb]{.906,  .902,  .902}$53.48$ & \cellcolor[rgb]{.906,  .902,  .902}$53.48$ & \cellcolor[rgb]{.906,  .902,  .902}$\textcolor{red}{\textbf{-0.00}}$ \\
    & \cellcolor[rgb]{.906,  .902,  .902}Qwen-72b & \cellcolor[rgb]{.906,  .902,  .902}$15.12$  & \cellcolor[rgb]{.906,  .902,  .902}$15.12$ & $\cellcolor[rgb]{.906,  .902,  .902}\textcolor{red}{\textbf{-0.00}}$ & \cellcolor[rgb]{.906,  .902,  .902}$53.88$ & \cellcolor[rgb]{.906,  .902,  .902}$53.88$ & \cellcolor[rgb]{.906,  .902,  .902}$\textcolor{red}{\textbf{-0.00}}$ & \cellcolor[rgb]{.906,  .902,  .902}$\textbf{54.66}$ & \cellcolor[rgb]{.906,  .902,  .902}$\textbf{54.66}$ & $\cellcolor[rgb]{.906,  .902,  .902}\textcolor{red}{\textbf{-0.00}}$ \\
    & ChatGPT & $\textbf{37.34}$  & $\textbf{37.34}$ & $\textcolor{red}{\textbf{-0.00}}$ & $\textbf{54.21}$ & $\textbf{54.21}$ & $\textcolor{red}{\textbf{-0.00}}$ & $54.45$ & $54.45$ & $\textcolor{red}{\textbf{-0.00}}$ \\
    \hdashline
    \multirow{7}[1]{*}{Specialized} 
    & GPT\_BigCode & $0.34$  & $0.34$ & $\textcolor{red}{\textbf{-0.00}}$ & $12.28$ & $12.28$ & $\textcolor{red}{\textbf{-0.00}}$ & $13.63$ & $13.63$ & $\textcolor{red}{\textbf{-0.00}}$ \\
    & \cellcolor[rgb]{.906,  .902,  .902}CodeLlama-34b & \cellcolor[rgb]{.906,  .902,  .902}$4.08$  & \cellcolor[rgb]{.906,  .902,  .902}$4.08$ & \cellcolor[rgb]{.906,  .902,  .902}$\textcolor{red}{\textbf{-0.00}}$ & \cellcolor[rgb]{.906,  .902,  .902}$47.36$ & \cellcolor[rgb]{.906,  .902,  .902}$47.36$ & \cellcolor[rgb]{.906,  .902,  .902}$\textcolor{red}{\textbf{-0.00}}$ & \cellcolor[rgb]{.906,  .902,  .902}$\textbf{48.99}$ & \cellcolor[rgb]{.906,  .902,  .902}$\textbf{48.99}$ & \cellcolor[rgb]{.906,  .902,  .902}$\textcolor{red}{\textbf{-0.00}}$ \\
    & CodeLlama-13b-Python & $5.31$  & $5.31$ & $\textcolor{red}{\textbf{-0.00}}$ & $44.37$ & $44.37$ & $\textcolor{red}{\textbf{-0.00}}$  & $46.39$ & $46.39$ & $\textcolor{red}{\textbf{-0.00}}$ \\
    & CodeLlama-7b & $6.38$  & $6.38$ & $\textcolor{red}{\textbf{-0.00}}$ & $38.44$ & $38.44$ & $\textcolor{red}{\textbf{-0.00}}$ & $40.02$ & $40.02$ & $\textcolor{red}{\textbf{-0.00}}$ \\
    & CodeLlama-7b-Python & $6.73$  & $6.73$ & $\textcolor{red}{\textbf{-0.00}}$ & $43.78$ & $43.78$ & $\textcolor{red}{\textbf{-0.00}}$ & $45.43$ & $45.43$ & $\textcolor{red}{\textbf{-0.00}}$ \\
    & StarCoder & $6.99$  & $6.99$ & $\textcolor{red}{\textbf{-0.00}}$ & $39.76$ & $39.76$ & $\textcolor{red}{\textbf{-0.00}}$ & $41.28$ & $41.28$ & $\textcolor{red}{\textbf{-0.00}}$ \\
    & CodeLlama-13b & $16.21$ & $16.21$ & $\textcolor{red}{\textbf{-0.00}}$ & $\textbf{47.72}$ & $\textbf{47.72}$ & $\textcolor{red}{\textbf{-0.00}}$ & $48.74$ & $48.74$  & $\textcolor{red}{\textbf{-0.00}}$ \\
    & WizardCoder-15b & $\textbf{16.79}$ & $\textbf{16.79}$ & $\textcolor{red}{\textbf{-0.00}}$ & $44.56$ & $44.56$ & $\textcolor{red}{\textbf{-0.00}}$ & $45.96$ & $45.96$ & $\textcolor{red}{\textbf{-0.00}}$ \\
    \bottomrule
    \end{tabular}%
    }
    \caption{\textbf{Scores of $23$ large language models (LLMs) on simple-level object-oriented programming (OOP) tasks}. We also reported the differences in evaluation results between \textit{pass@k} and \textit{pass@o}. (All LLMs are evaluated in zero-shot fashion. For \textit{pass@$100$} and \textit{pass@$80$} scores, we use a temperature of $0.8$ and top-$p$=$0.95$. For \textit{pass@$1$} scores, we use a temperature of $0.1$ and top-$p$=$0.95$. \sethlcolor{red}\hl{Red} indicates the differences evaluated using the \textit{pass@o} and \textit{pass@k} metrics; \underline{Underlined} indicates the maximum disparities evaluated between \textit{pass@o} and \textit{pass@k} metrics; \sethlcolor{gray}\hl{Gray} indicates models with a larger number of parameters.)}
  \label{tab:simple_level_OOP_scores}%
\end{table*}%

\begin{table*}[!t]
  \centering
    \resizebox{1.\linewidth}{!}{
    \begin{tabular}{clccccccccc}
    \toprule
    \multicolumn{2}{c}{\multirow{2}[2]{*}{\textbf{Model}}} & \multicolumn{3}{c}{\textbf{1}} & \multicolumn{3}{c}{\textbf{80}} & \multicolumn{3}{c}{\textbf{100}} \\
    \multicolumn{2}{c}{} & \textit{pass@$k$} & \textit{pass@$o$} & $\boldsymbol{\Delta}\left(\downarrow\right)$ & \textit{pass@$k$} & \textit{pass@$o$} & $\boldsymbol{\Delta}\left(\downarrow\right)$ & \textit{pass@$k$} & \textit{pass@$o$} & $\boldsymbol{\Delta}\left(\downarrow\right)$\\
    \midrule
    \multirow{15}[1]{*}{General}
    & Falcon-7b & $0.02$	& $0.00$ & $\textcolor{red}{\textbf{-0.02}}$ & $0.22$	& $0.00$ & $\textcolor{red}{\textbf{-0.22}}$ & $0.28$ & $0.00$ & $\textcolor{red}{\textbf{-0.28}}$ \\
    & Falcon-40b & $0.02$	& $0.00$ & $\textcolor{red}{\textbf{-0.02}}$ & $0.23$	& $0.00$ & $\textcolor{red}{\textbf{-0.23}}$ & $0.72$ & $0.00$ & $\textcolor{red}{\textbf{-0.72}}$\\
    & Llama2-7b & $0.02$	& $0.00$ & $\textcolor{red}{\textbf{-0.02}}$ & $5.51$	& $0.00$ & $\textcolor{red}{\textbf{-5.51}}$ & $6.41$ & $0.00$ & $\textcolor{red}{\textbf{-6.41}}$ \\
    & InternLm-7b & $0.03$  & $0.00$ & $\textcolor{red}{\textbf{-0.03}}$ & $1.03$  & $0.00$ & $\textcolor{red}{\textbf{-1.03}}$ & $1.26$  & $0.00$ & $\textcolor{red}{\textbf{-1.26}}$ \\
    & Llama2-13b & $0.08$  & $0.00$ & $\textcolor{red}{\textbf{-0.08}}$ & $11.78$  & $0.00$ & $\textcolor{red}{\textbf{-11.78}}$ & $13.39$  & $0.00$ & $\textcolor{red}{\textbf{-13.39}}$ \\
    & Yi-6b & $0.08$  & $0.00$ & $\textcolor{red}{\textbf{-0.08}}$ & $6.23$  & $0.36$ & $\textcolor{red}{\textbf{-5.87}}$ & $7.39$  & $0.42$ & $\textcolor{red}{\textbf{-6.97}}$ \\
    & MPT-7b & $0.61$  & $0.00$ & $\textcolor{red}{\textbf{-0.61}}$  & $8.16$  & $0.00$ & $\textcolor{red}{\textbf{-8.16}}$ & $9.38$  & $0.00$ & $\textcolor{red}{\textbf{-9.38}}$ \\
    & Qwen-7b & $0.80$  & $0.00$ & $\textcolor{red}{\textbf{-0.80}}$ & $20.79$  & $0.00$ & $\textcolor{red}{\textbf{-20.79}}$ & $23.27$  & $0.00$ & $\textcolor{red}{\textbf{-23.27}}$ \\
    & DeepSeek-7b & $1.51$  & $0.00$ & $\textcolor{red}{\textbf{-1.51}}$ & $15.47$  & $0.45$ & $\textcolor{red}{\textbf{-15.02}}$ & $17.14$  & $0.56$ & $\textcolor{red}{\textbf{-16.58}}$ \\
    & Qwen-14b & $1.82$  & $0.00$ & $\textcolor{red}{\textbf{-1.82}}$ & $37.58$  & $5.12$ & $\textcolor{red}{\textbf{-32.46}}$ & $40.10$  & $6.12$ & $\textcolor{red}{\textbf{-33.98}}$ \\
    & Yi-34b & $2.10$  & $0.00$ & $\textcolor{red}{\textbf{-2.10}}$ & $25.61$  & $0.58$ & $\textcolor{red}{\textbf{-25.03}}$ & $27.79$  & $0.70$ & $\textcolor{red}{\textbf{-27.09}}$ \\
    & \cellcolor[rgb]{.906,  .902,  .902}Llama2-70b & \cellcolor[rgb]{.906,  .902,  .902}$5.01$  & \cellcolor[rgb]{.906,  .902,  .902}$0.00$ & $\cellcolor[rgb]{.906,  .902,  .902}\textcolor{red}{\textbf{-5.01}}$ & \cellcolor[rgb]{.906,  .902,  .902}$21.94$  & \cellcolor[rgb]{.906,  .902,  .902}$1.34$ & \cellcolor[rgb]{.906,  .902,  .902}$\textcolor{red}{\textbf{-20.60}}$ & \cellcolor[rgb]{.906,  .902,  .902}$24.27$  & \cellcolor[rgb]{.906,  .902,  .902}$1.68$ & \cellcolor[rgb]{.906,  .902,  .902}$\textcolor{red}{\textbf{-22.59}}$ \\
    & \cellcolor[rgb]{.906,  .902,  .902}DeepSeek-67b & \cellcolor[rgb]{.906,  .902,  .902}$7.89$  & \cellcolor[rgb]{.906,  .902,  .902}$0.00$ & $\cellcolor[rgb]{.906,  .902,  .902}\textcolor{red}{\textbf{-7.89}}$ & \cellcolor[rgb]{.906,  .902,  .902}$49.79$  & \cellcolor[rgb]{.906,  .902,  .902}$13.03$ & \cellcolor[rgb]{.906,  .902,  .902}$\textcolor{red}{\textbf{\underline{-36.76}}}$ & \cellcolor[rgb]{.906,  .902,  .902}$52.43$  & \cellcolor[rgb]{.906,  .902,  .902}$15.30$ & \cellcolor[rgb]{.906,  .902,  .902}$\textcolor{red}{\textbf{\underline{-37.13}}}$ \\
    & \cellcolor[rgb]{.906,  .902,  .902}Qwen-72b & \cellcolor[rgb]{.906,  .902,  .902}$13.02$  & \cellcolor[rgb]{.906,  .902,  .902}$0.28$ & $\cellcolor[rgb]{.906,  .902,  .902}\textcolor{red}{\textbf{-12.74}}$  & \cellcolor[rgb]{.906,  .902,  .902}$63.41$  & \cellcolor[rgb]{.906,  .902,  .902}$26.97$ & \cellcolor[rgb]{.906,  .902,  .902}$\textcolor{red}{\textbf{-36.44}}$ & \cellcolor[rgb]{.906,  .902,  .902}$65.21$  & \cellcolor[rgb]{.906,  .902,  .902}$29.71$ & \cellcolor[rgb]{.906,  .902,  .902}$\textcolor{red}{\textbf{-35.50}}$ \\
    & ChatGPT & $\textbf{51.71}$  & $\textbf{2.53}$ & $\textcolor{red}{\textbf{\underline{-49.18}}}$ & $\textbf{83.30}$ & $\textbf{51.54}$ & $\textcolor{red}{\textbf{-31.76}}$ & $\textbf{83.67}$ & $\textbf{54.78}$ & $\textcolor{red}{\textbf{-28.89}}$\\
    \hdashline
    \multirow{7}[1]{*}{Specialized} 
    & GPT\_BigCode & $0.08$  & $0.00$ & $\textcolor{red}{\textbf{-0.08}}$ & $9.22$  & $0.67$ & $\textcolor{red}{\textbf{-8.55}}$ & $10.55$  & $0.84$ & $\textcolor{red}{\textbf{-9.71}}$ \\
    & CodeLlama-7b & $3.46$  & $0.00$ & $\textcolor{red}{\textbf{-3.46}}$ & $36.85$  & $3.66$ & $\textcolor{red}{\textbf{-33.19}}$ & $39.15$  & $4.40$ & $\textcolor{red}{\textbf{-34.75}}$ \\
    & CodeLlama-13b-Python & $4.31$  & $0.01$ & $\textcolor{red}{\textbf{-4.30}}$ & $42.12$  & $10.06$ & $\textcolor{red}{\textbf{-32.06}}$ & $45.07$  & $11.84$ & $\textcolor{red}{\textbf{-33.23}}$ \\
    & StarCoder & $8.01$  & $0.01$ & $\textcolor{red}{\textbf{-8.00}}$ & $44.40$  & $4.28$ & $\textcolor{red}{\textbf{-40.12}}$ & $46.70$  & $5.07$ & $\textcolor{red}{\textbf{-41.63}}$ \\
    & CodeLlama-7b-Python & $8.13$  & $0.01$ & $\textcolor{red}{\textbf{-8.12}}$ & $43.96$  & $9.28$ & $\textcolor{red}{\textbf{-34.68}}$ & $46.17$  & $10.76$ & $\textcolor{red}{\textbf{-35.41}}$ \\
    & WizardCoder-15b & $9.10$  & $0.00$ & $\textcolor{red}{\textbf{-9.10}}$ & $45.25$  & $1.29$ & $\textcolor{red}{\textbf{-43.96}}$ & $47.41$  & $1.50$ & $\textcolor{red}{\textbf{\underline{-45.91}}}$ \\
    & CodeLlama-13b & $9.46$  & $0.00$ & $\textcolor{red}{\textbf{-9.46}}$ & $\textbf{51.73}$  & $7.62$ & $\textcolor{red}{\textbf{\underline{-44.11}}}$ & $\textbf{54.55}$  & $9.12$ & $\textcolor{red}{\textbf{-45.43}}$ \\
    & \cellcolor[rgb]{.906,  .902,  .902}CodeLlama-34b & $\cellcolor[rgb]{.906,  .902,  .902}\textbf{10.23}$  & $\cellcolor[rgb]{.906,  .902,  .902}0.00$ & $\cellcolor[rgb]{.906,  .902,  .902}\textcolor{red}{\textbf{\underline{-10.23}}}$ & \cellcolor[rgb]{.906,  .902,  .902}$51.68$  & \cellcolor[rgb]{.906,  .902,  .902}$\textbf{11.41}$ & \cellcolor[rgb]{.906,  .902,  .902}$\textcolor{red}{\textbf{-40.27}}$ & \cellcolor[rgb]{.906,  .902,  .902}$54.22$  & \cellcolor[rgb]{.906,  .902,  .902}$\textbf{13.48}$ & \cellcolor[rgb]{.906,  .902,  .902}$\textcolor{red}{\textbf{-40.74}}$ \\
    \bottomrule
    \end{tabular}%
    }
    \caption{\textbf{Scores of $23$ large language models (LLMs) on moderate-level object-oriented programming (OOP) tasks}. We also reported the differences in evaluation results between \textit{pass@k} and \textit{pass@o}. (All LLMs are evaluated in zero-shot fashion. For \textit{pass@$100$} and \textit{pass@$80$} scores, we use a temperature of $0.8$ and top-$p$=$0.95$. For \textit{pass@$1$} scores, we use a temperature of $0.1$ and top-$p$=$0.95$. \sethlcolor{red}\hl{Red} indicates the differences evaluated using the \textit{pass@o} and \textit{pass@k} metrics; \underline{Underlined} indicates the maximum disparities evaluated between \textit{pass@o} and \textit{pass@k} metrics; \sethlcolor{gray}\hl{Gray} indicates models with a larger number of parameters.)}
  \label{tab:moderate_level_OOP_scores}%
\end{table*}%

\begin{table*}[!t]
  \centering
    \resizebox{1.\linewidth}{!}{
    \begin{tabular}{clccccccccc}
    \toprule
    \multicolumn{2}{c}{\multirow{2}[2]{*}{\textbf{Model}}} & \multicolumn{3}{c}{\textbf{1}} & \multicolumn{3}{c}{\textbf{80}} & \multicolumn{3}{c}{\textbf{100}} \\
    \multicolumn{2}{c}{} & \textit{pass@$k$} & \textit{pass@$o$} & $\boldsymbol{\Delta}\left(\downarrow\right)$ & \textit{pass@$k$} & \textit{pass@$o$} & $\boldsymbol{\Delta}\left(\downarrow\right)$ & \textit{pass@$k$} & \textit{pass@$o$} & $\boldsymbol{\Delta}\left(\downarrow\right)$\\
    \midrule
    \multirow{15}[1]{*}{General}
    & Llama2-7b & $0.00$ & $0.00$ & $\textcolor{red}{\textbf{-0.00}}$ & $0.00$ & $0.00$ & $\textcolor{red}{\textbf{-0.00}}$ & $0.00$ & $0.00$ & $\textcolor{red}{\textbf{-0.00}}$ \\
    & Falcon-7b & $0.00$ & $0.00$ & $\textcolor{red}{\textbf{-0.00}}$ & $0.22$ & $0.00$ & $\textcolor{red}{\textbf{-0.22}}$ & $0.28$ & $0.00$ & $\textcolor{red}{\textbf{-0.28}}$ \\
    & MPT-7b & $0.00$  & $0.00$ & $\textcolor{red}{\textbf{-0.00}}$ & $0.23$  & $0.00$ & $\textcolor{red}{\textbf{-0.23}}$ & $0.29$  & $0.00$ & $\textcolor{red}{\textbf{-0.29}}$ \\
    & Llama2-13b & $0.00$  & $0.00$ & $\textcolor{red}{\textbf{-0.00}}$ & $0.47$  & $0.00$ & $\textcolor{red}{\textbf{-0.47}}$ & $0.58$  & $0.00$ & $\textcolor{red}{\textbf{-0.58}}$ \\
    & Qwen-7b & $0.01$  & $0.01$ & $\textcolor{red}{\textbf{-0.00}}$ & $2.08$  & $0.46$ & $\textcolor{red}{\textbf{-1.62}}$ & $2.47$  & $0.51$ & $\textcolor{red}{\textbf{-1.96}}$ \\
    & InternLm-7b & $0.01$  & $0.00$ & $\textcolor{red}{\textbf{-0.01}}$ & $0.23$  & $0.00$ & $\textcolor{red}{\textbf{-0.23}}$ & $0.29$  & $0.00$ &$\textcolor{red}{\textbf{-0.29}}$\\
    & Falcon-40b & $0.02$  & $0.00$ & $\textcolor{red}{\textbf{-0.01}}$ & $0.23$  & $0.00$ & $\textcolor{red}{\textbf{-0.23}}$ & $0.72$  & $0.00$ & $\textcolor{red}{\textbf{-0.72}}$\\
    & Qwen-14b & $0.07$  & $0.06$ & $\textcolor{red}{\textbf{-0.01}}$ & $9.77$  & $4.70$ & $\textcolor{red}{\textbf{-5.07}}$ & $11.24$  & $5.53$ & $\textcolor{red}{\textbf{-5.73}}$ \\
    & Yi-6b & $0.09$  & $0.03$ & $\textcolor{red}{\textbf{-0.06}}$ & $3.52$  & $1.16$ & $\textcolor{red}{\textbf{-2.36}}$ & $4.20$  & $1.45$ & $\textcolor{red}{\textbf{-2.75}}$ \\
    & DeepSeek-7b & $1.51$  & $0.00$ & $\textcolor{red}{\textbf{-1.51}}$ & $15.47$  & $0.45$ & $\textcolor{red}{\textbf{-15.02}}$ & $17.14$  & $0.56$ & $\textcolor{red}{\textbf{-16.58}}$\\
    & Yi-34b & $1.77$  & $1.20$ & $\textcolor{red}{\textbf{-0.57}}$ & $16.27$  & $8.66$ & $\textcolor{red}{\textbf{-7.61}}$ & $17.62$  & $9.83$ & $\textcolor{red}{\textbf{-7.79}}$\\
    & \cellcolor[rgb]{.906,  .902,  .902}Llama2-70b & \cellcolor[rgb]{.906,  .902,  .902}$1.94$  & \cellcolor[rgb]{.906,  .902,  .902}$1.42$ & \cellcolor[rgb]{.906,  .902,  .902}$\textcolor{red}{\textbf{-0.52}}$ & \cellcolor[rgb]{.906,  .902,  .902}$17.21$  & \cellcolor[rgb]{.906,  .902,  .902}$11.26$ & \cellcolor[rgb]{.906,  .902,  .902}$\textcolor{red}{\textbf{-5.95}}$ & \cellcolor[rgb]{.906,  .902,  .902}$19.05$  & \cellcolor[rgb]{.906,  .902,  .902}$12.81$ & \cellcolor[rgb]{.906,  .902,  .902}$\textcolor{red}{\textbf{-6.24}}$\\
    & \cellcolor[rgb]{.906,  .902,  .902}Qwen-72b & \cellcolor[rgb]{.906,  .902,  .902}$7.54$  & \cellcolor[rgb]{.906,  .902,  .902}$4.43$ & $\cellcolor[rgb]{.906,  .902,  .902}\textcolor{red}{\textbf{-3.11}}$ & \cellcolor[rgb]{.906,  .902,  .902}$53.28$  & \cellcolor[rgb]{.906,  .902,  .902}$36.56$ & \cellcolor[rgb]{.906,  .902,  .902}$\textcolor{red}{\textbf{-16.72}}$ & \cellcolor[rgb]{.906,  .902,  .902}$56.11$  & \cellcolor[rgb]{.906,  .902,  .902}$38.67$ & \cellcolor[rgb]{.906,  .902,  .902}$\textcolor{red}{\textbf{-17.44}}$\\
    & \cellcolor[rgb]{.906,  .902,  .902}DeepSeek-67b & \cellcolor[rgb]{.906,  .902,  .902}$7.89$  & \cellcolor[rgb]{.906,  .902,  .902}$0.00$ & $\cellcolor[rgb]{.906,  .902,  .902}\textcolor{red}{\textbf{-7.89}}$ & \cellcolor[rgb]{.906,  .902,  .902}$49.79$  & \cellcolor[rgb]{.906,  .902,  .902}$13.03$ & \cellcolor[rgb]{.906,  .902,  .902}$\textcolor{red}{\textbf{\underline{-36.76}}}$ & \cellcolor[rgb]{.906,  .902,  .902}$52.43$  & \cellcolor[rgb]{.906,  .902,  .902}$15.30$ & \cellcolor[rgb]{.906,  .902,  .902}$\textcolor{red}{\textbf{\underline{-37.13}}}$\\
    & ChatGPT & $\textbf{36.52}$  & $\textbf{19.70}$ & $\textcolor{red}{\textbf{\underline{-16.82}}}$ & $\textbf{78.94}$ & $\textbf{71.83}$ & $\textcolor{red}{\textbf{-7.11}}$ & $\textbf{79.95}$ & $\textbf{73.37}$ & $\textcolor{red}{\textbf{-6.58}}$\\
    \hdashline
    \multirow{7}[1]{*}{Specialized} 
    & WizardCoder-15b & $0.00$  & $0.00$ & $\textcolor{red}{\textbf{-0.00}}$ & $3.05$  & $2.35$ & $\textcolor{red}{\textbf{-0.70}}$ & $3.48$  & $2.76$ & $\textcolor{red}{\textbf{-0.72}}$\\
    & CodeLlama-13b & $0.00$  & $0.00$ & $\textcolor{red}{\textbf{-0.00}}$ & $5.69$  & $1.07$ & $\textcolor{red}{\textbf{-4.62}}$ & $6.75$  & $1.31$ & $\textcolor{red}{\textbf{-5.44}}$\\
    & StarCoder & $0.00$  & $0.00$ & $\textcolor{red}{\textbf{-0.00}}$ & $7.34$  & $2.74$ & $\textcolor{red}{\textbf{-4.60}}$ & $8.65$  & $3.31$ & $\textcolor{red}{\textbf{-5.34}}$\\
    & GPT\_BigCode & $0.01$  & $0.00$ & $\textcolor{red}{\textbf{-0.01}}$ & $2.08$  & $0.23$ & $\textcolor{red}{\textbf{-1.85}}$ & $2.54$  & $0.29$ & $\textcolor{red}{\textbf{-2.25}}$\\
    & CodeLlama-7b-Python & $0.17$  & $0.15$ & $\textcolor{red}{\textbf{-0.02}}$ & $5.93$  & $2.84$ & $\textcolor{red}{\textbf{-3.09}}$ & $7.02$  & $3.49$ & $\textcolor{red}{\textbf{-3.53}}$\\
    & CodeLlama-13b-Python & $0.17$  & $0.17$ & $\textcolor{red}{\textbf{-0.00}}$ & $26.74$  & $16.61$ & $\textcolor{red}{\textbf{-10.13}}$ & $25.75$  & $18.64$ & $\textcolor{red}{\textbf{-7.11}}$\\
    & CodeLlama-7b & $0.18$  & $0.13$ & $\textcolor{red}{\textbf{-0.05}}$ & $4.65$  & $1.77$ & $\textcolor{red}{\textbf{-2.88}}$ & $5.64$  & $2.18$ & $\textcolor{red}{\textbf{-3.46}}$ \\
    & \cellcolor[rgb]{.906,  .902,  .902}CodeLlama-34b & \cellcolor[rgb]{.906,  .902,  .902}$\textbf{3.08}$  & \cellcolor[rgb]{.906,  .902,  .902}$\textbf{2.13}$ & \cellcolor[rgb]{.906,  .902,  .902}$\textcolor{red}{\textbf{\underline{-0.95}}}$ & \cellcolor[rgb]{.906,  .902,  .902}$\textbf{40.26}$  & \cellcolor[rgb]{.906,  .902,  .902}$\textbf{27.83}$ & \cellcolor[rgb]{.906,  .902,  .902}$\textcolor{red}{\textbf{\underline{-12.43}}}$ & \cellcolor[rgb]{.906,  .902,  .902}$\textbf{43.59}$  & \cellcolor[rgb]{.906,  .902,  .902}$\textbf{30.51}$ & \cellcolor[rgb]{.906,  .902,  .902}$\textcolor{red}{\textbf{\underline{-13.08}}}$ \\
    \bottomrule
    \end{tabular}%
    }
    \caption{\textbf{Scores of $23$ large language models (LLMs) on difficult-level object-oriented programming (OOP) tasks}. We also reported the differences in evaluation results between \textit{pass@k} and \textit{pass@o}. (All LLMs are evaluated in zero-shot fashion. For \textit{pass@$100$} and \textit{pass@$80$} scores, we use a temperature of $0.8$ and top-$p$=$0.95$. For \textit{pass@$1$} scores, we use a temperature of $0.1$ and top-$p$=$0.95$. \sethlcolor{red}\hl{Red} indicates the differences evaluated using the \textit{pass@o} and \textit{pass@k} metrics; \underline{Underlined} indicates the maximum disparities evaluated between \textit{pass@o} and \textit{pass@k} metrics; \sethlcolor{gray}\hl{Gray} indicates models with a larger number of parameters.)}
  \label{tab:difficult_level_OOP_scores}%
\end{table*}%

\begin{figure}[!t]
    \centering
    \includegraphics[width=0.48\textwidth]{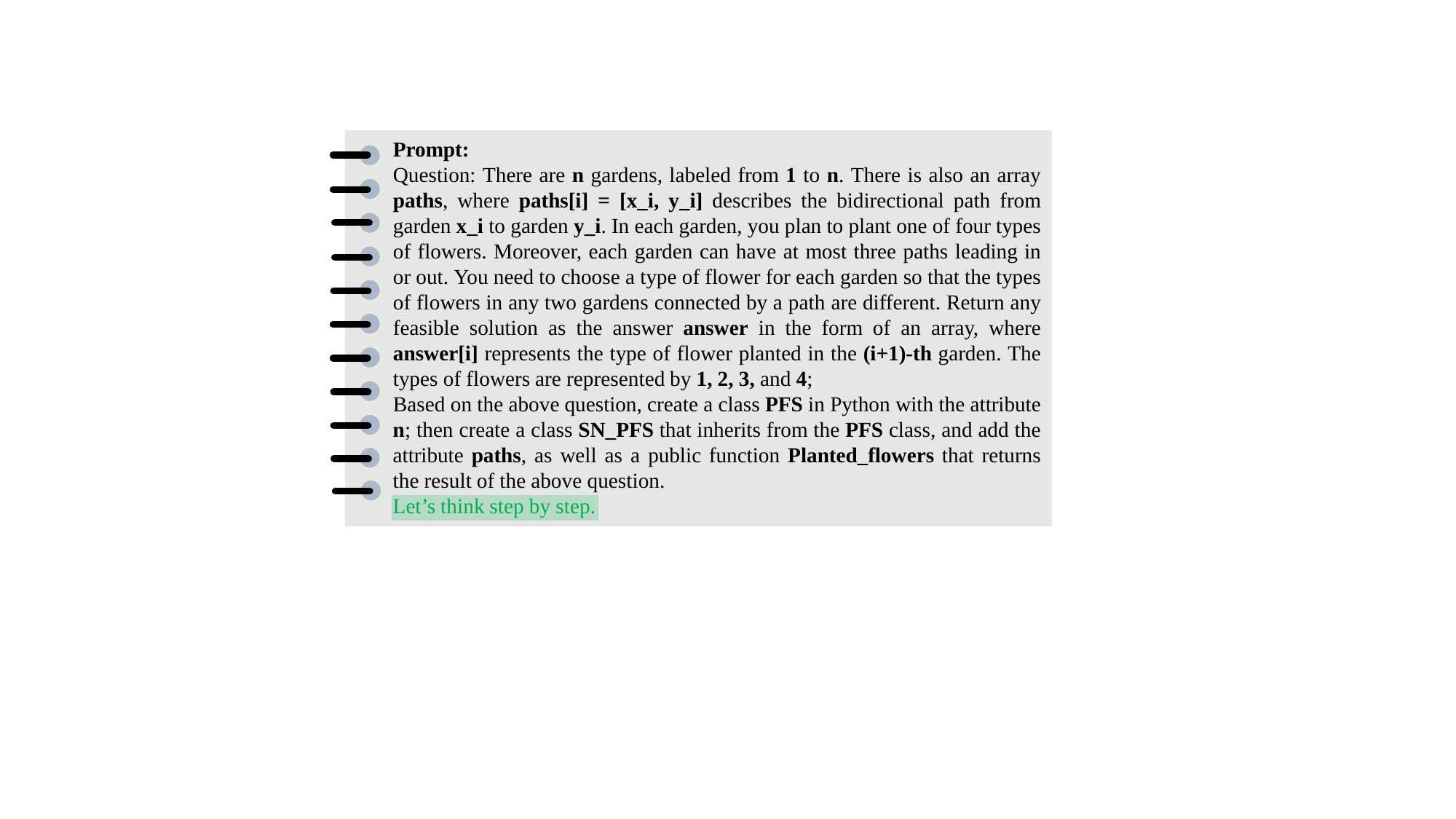}
    \caption{\textbf{Example of a prompt using the zero-shot CoT approach}. (The green content indicates guiding the model to generate code step by step using the CoT approach.)}
    \label{fig:zero_shot_cot}
\end{figure}

\begin{figure}[!t]
    \centering
    \includegraphics[width=0.48\textwidth]{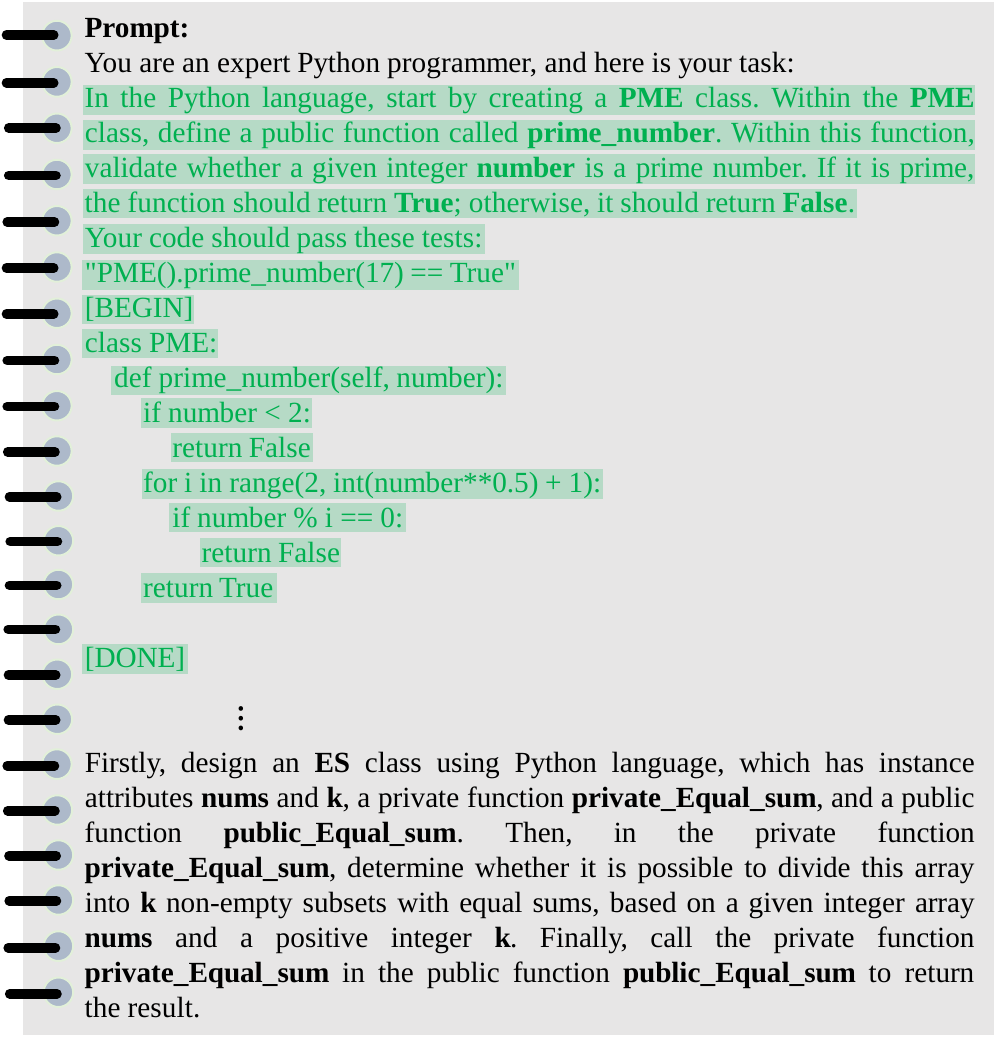}
    \caption{\textbf{Prompt using the few-shot approach.} (The  \sethlcolor{green}\hl{green color} indicates the added few-shot content.)}
    \label{fig:few_shot}
\end{figure}

\begin{figure*}[!t]
    \centering
    \includegraphics[width=\textwidth]{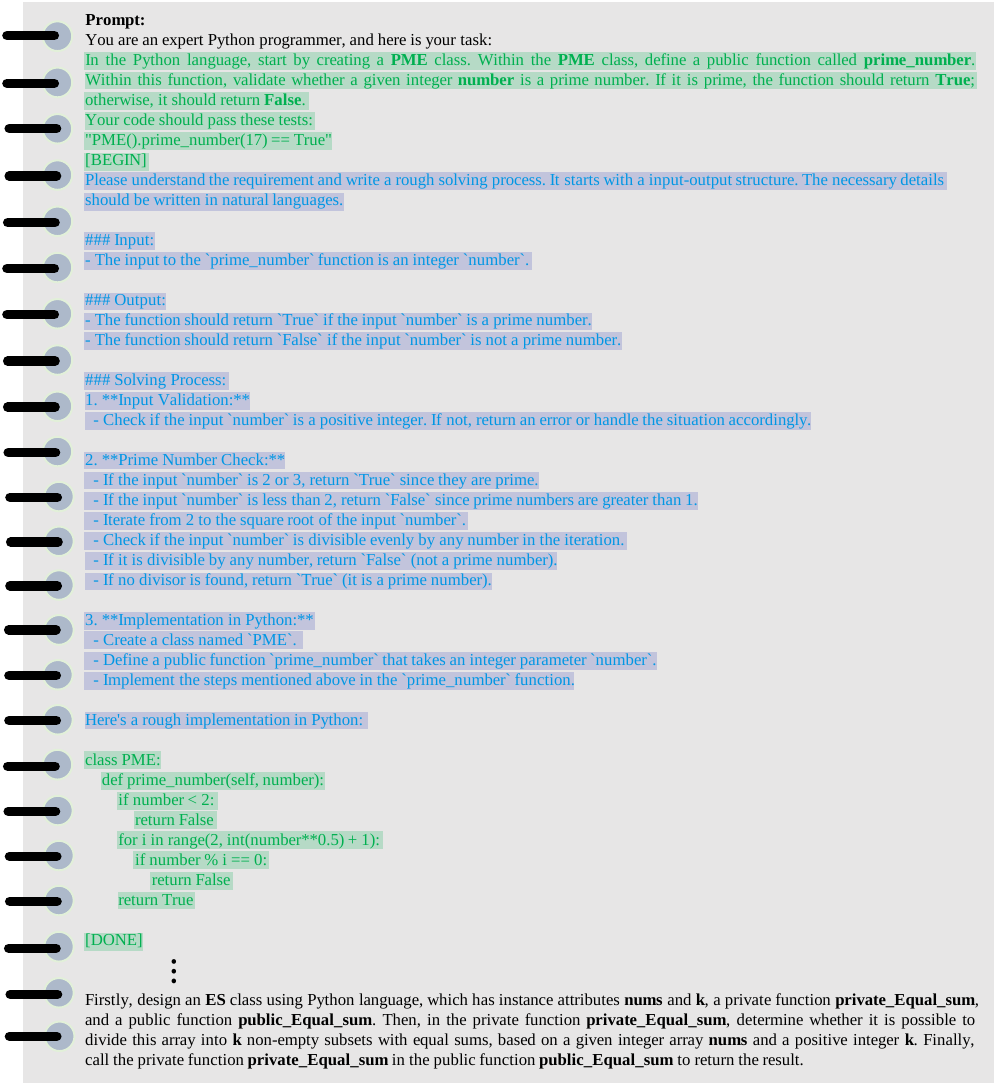}
    \caption{\textbf{Prompt of using the few-shot CoT approach.} (The \sethlcolor{green}\hl{green color} indicates the added few-shot content; The \sethlcolor{blue}\hl{blue color} indicates guiding the model to generate code step by step using the CoT approach.)}
    \label{fig:few_shot_shot}
\end{figure*}

\begin{figure*}[!t]
    \centering
    \includegraphics[width=\textwidth]{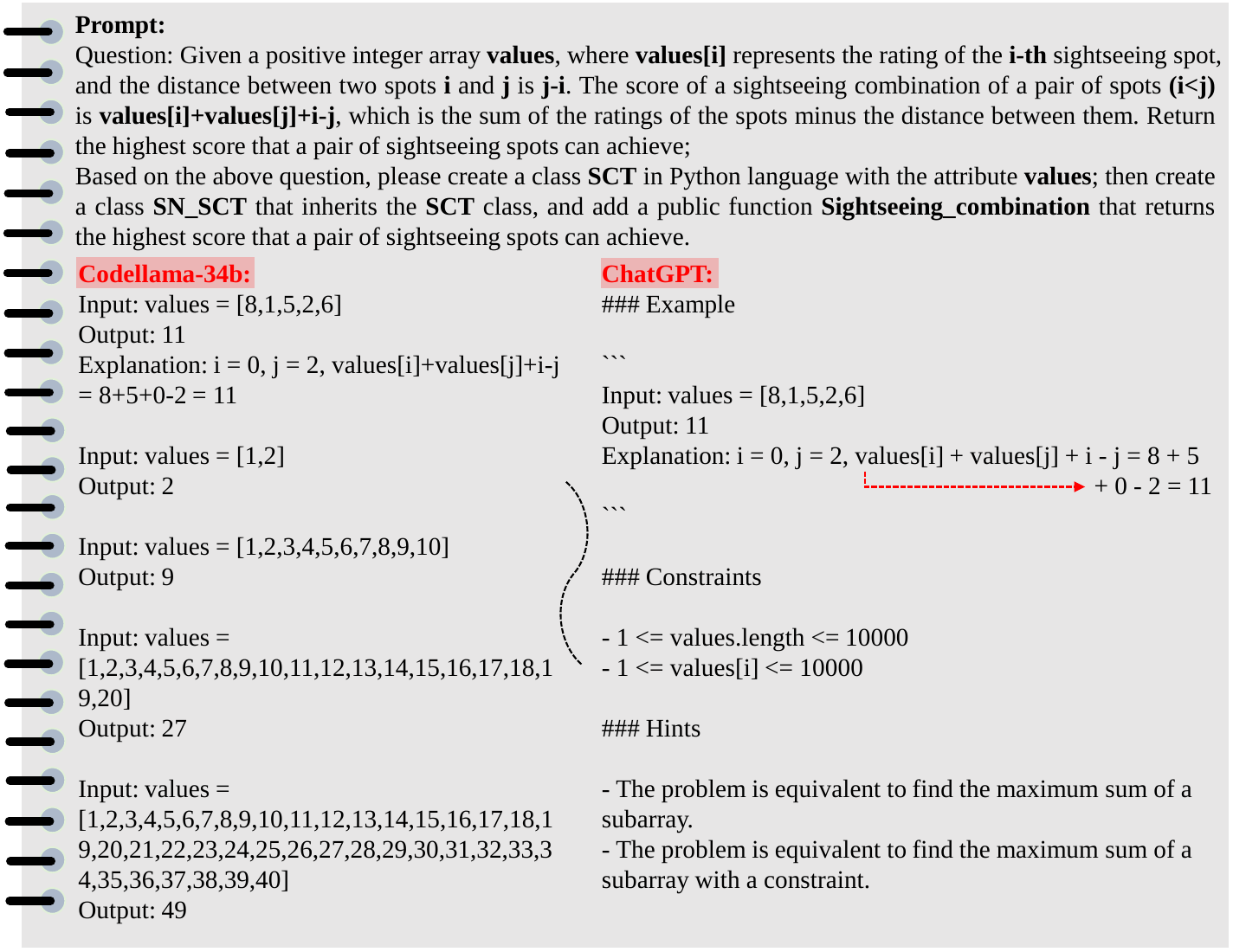}
    \caption{\textbf{An example of Code generated by CodeLlama-34b and ChatGPT.} We can see that CodeLlama-34b did not generate the corresponding class and public function.}
    \label{fig:CodeLlama34b_vs_ChatGPT_01}
\end{figure*}

\begin{figure*}[!t]
    \centering
    \includegraphics[width=\textwidth]{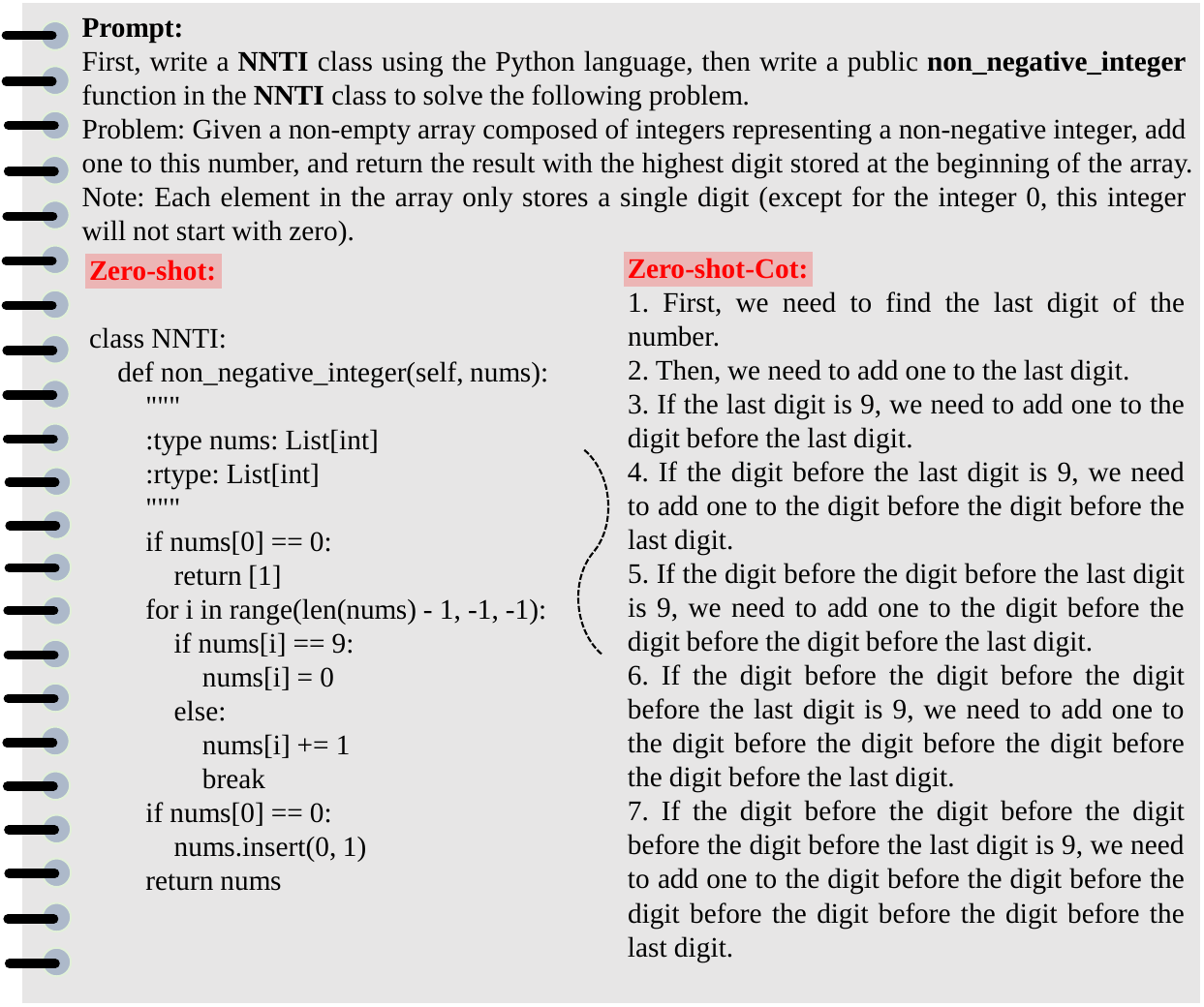}
    \caption{\textbf{Comparison of results generated by zero-shot and zero-shot CoT}. We can see that: 1) using the zero-shot CoT approach can lead the model to generate illusions, thus preventing it from generating the corresponding code. 2) using the zero-shot approach, the model is directly prompted to generate the corresponding code.}
    \label{fig:zero_shot_vs_zero_shot_cot}
\end{figure*}

\begin{figure*}[!t]
    \centering
    \includegraphics[width=0.8\textwidth]{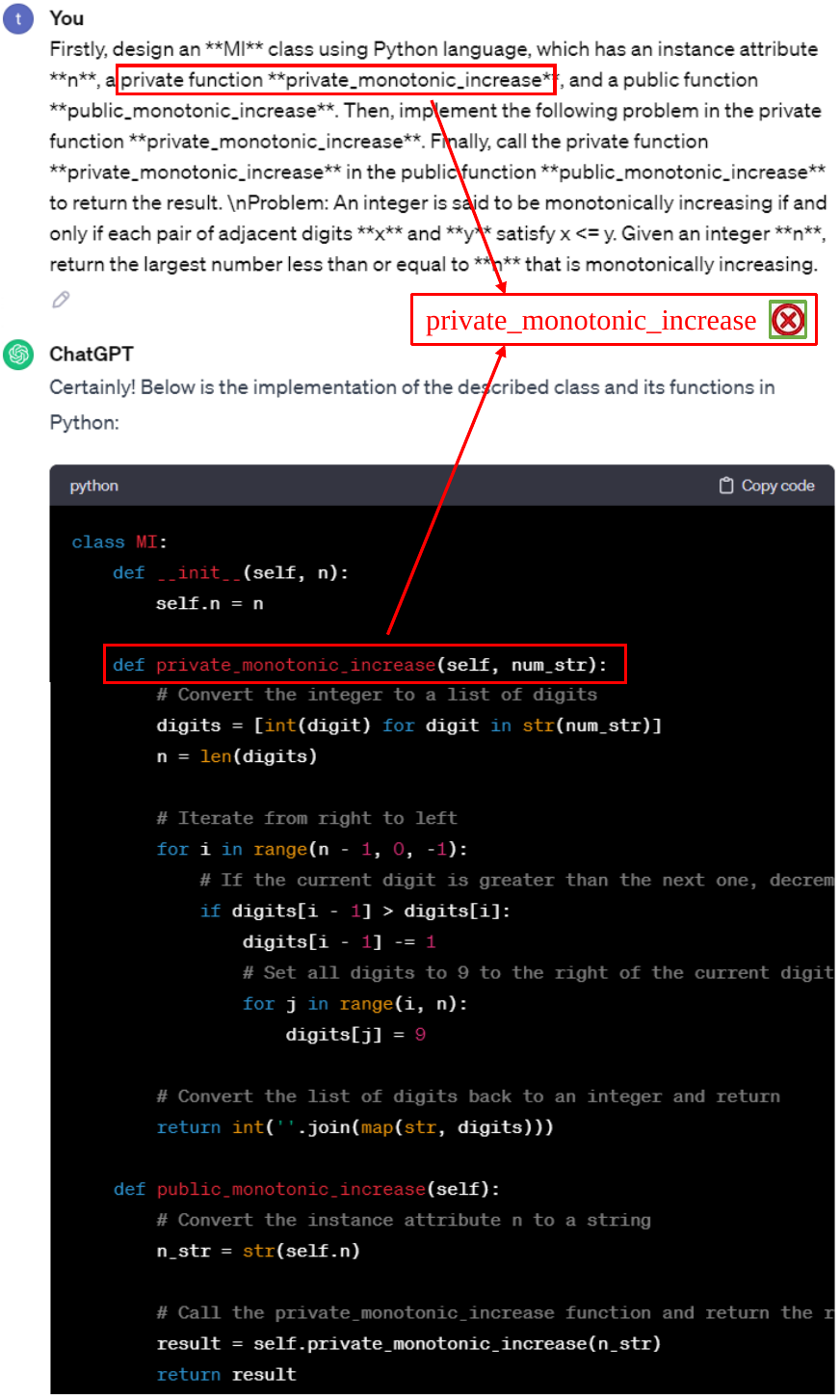}
    \caption{\textbf{Case 1 of generating code using the web version of ChatGPT}.}
    \label{fig:Web_ChatGPT_code01}
\end{figure*}

\begin{figure*}[!t]
    \centering
    \includegraphics[width=0.8\textwidth]{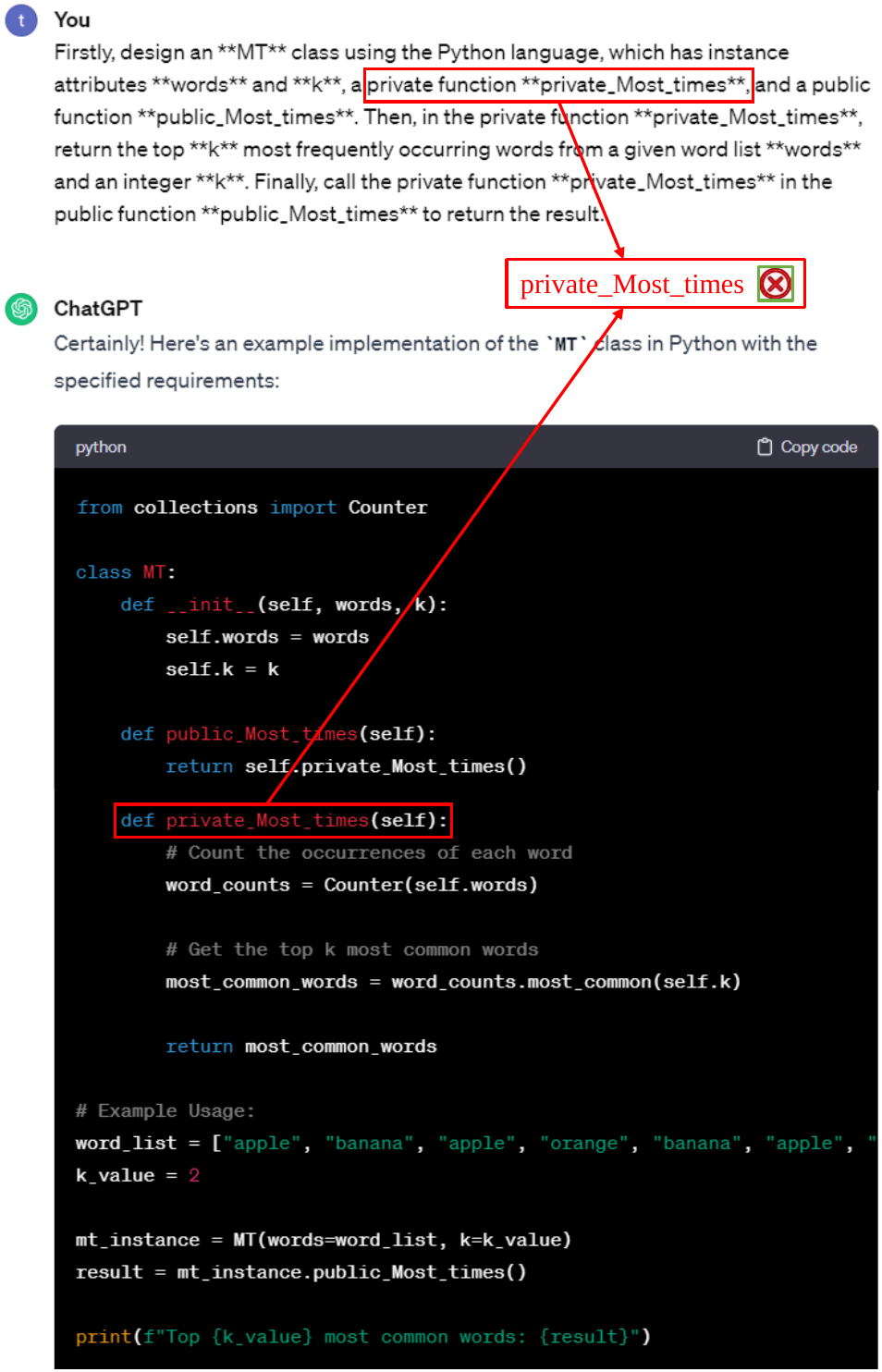}
    \caption{\textbf{Case 2 of generating code using the web version of ChatGPT}.}
    \label{fig:Web_ChatGPT_code02}
\end{figure*}

\begin{figure*}[!t]
    \centering
    \includegraphics[width=0.8\textwidth]{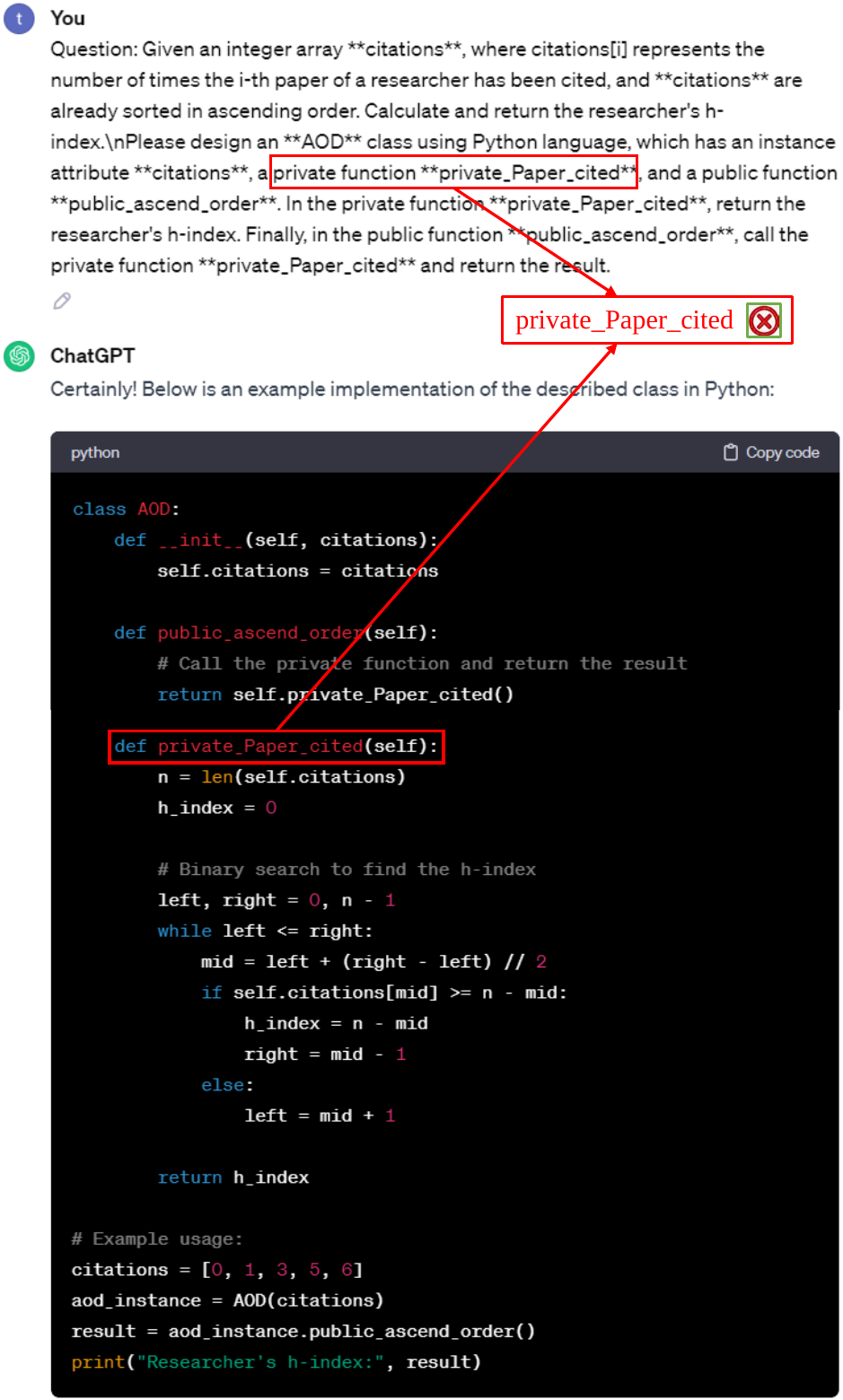}
    \caption{\textbf{Case 3 of generating code using the web version of ChatGPT}.}
    \label{fig:Web_ChatGPT_code03}
\end{figure*}

\section{\textit{Pass@$k$} calculation process}
\label{sec:appendix_related_work}
The calculation process for \textit{pass@$k$} is:
\begin{equation}
\label{eq:mertric_pass@k}
\textit{pass@$k$}:=\mathop{\mathbb{E}}_{Problems} \left[1-\frac{{\binom{n-c}{k}}}{\binom{n}{k}} \right]
\end{equation}

In Eq. (\ref{eq:mertric_pass@k}), $n$ represents the number of code generations for a given problem; $c$ represents the quantity of $n$ generated codes passing tests.

\section{Limitations of HumanEval and MBPP benchmarks}
\label{sec:appendix_motivate}
Existing HumanEval~\cite{chen2021evaluating} and MBPP~\cite{austin2021program} benchmarks primarily focus on FP to evaluate the programming capabilities of LLMs, as illustrated in Figure~\ref{fig:humaneval_and_mbpp_shortcoming}. 


\section{Detailed construction process of OOP.}
\label{sec:appendix_level_classification}
In the process of establishing the OOP benchmark, we hired a total of nine fourth-year undergraduate computer science students. Among them, two students were involved in the data collection process, four students participated in the rewriting process, and two students contributed to the use case construction phase, as shown in Figure~\ref{fig:oop_construction}.

During the data collection process, problems or requirements described in Non-English natural language are translated using the Google API, followed by manual verification. In the use case construction phase, we begin by inputting the rewritten prompt into ChatGPT to generate the corresponding code. Subsequently, the generated code is used for input testing. Finally, the output results are saved along with the input tests to serve as test cases. However, the code generated by ChatGPT may not always be correct, requiring manual inspection and correction. During the process of building the benchmark for OOP, we spent a total of \$200.

\begin{table}[!t]
  \centering
  \resizebox{\linewidth}{!}{
    \begin{tabular}{lcccc}
    \toprule
    \textbf{Model name} & \textbf{Size}  & \multicolumn{1}{c}{\textbf{Years}} & \multicolumn{1}{c}{\textbf{Open-source}} & \multicolumn{1}{c}{\textbf{Task type}} \\
    \midrule
    Falcon & 7b, 40b & $2023$  & \CheckmarkBold & General \\
    DeepSeek & 7b, 67b & $2023$  & \CheckmarkBold & General \\
    Llama2 & 7b, 13b, 70b & $2023$  & \CheckmarkBold & General \\
    Yi & 6b, 34b & $2023$  & \CheckmarkBold & General \\
    InternLm & 7b    & $2023$  & \CheckmarkBold & General \\
    MPT   & 7b    & $2023$  & \CheckmarkBold & General \\
    Qwen  & 7b, 14b, 72b & $2023$  & \CheckmarkBold & General \\
    ChatGPT & N/A   & $2023$  & \XSolidBrush  & General \\
    GPT\_BigCode & 1.12b & $2023$  & \CheckmarkBold & Code-specialized\\
    CodeLlama & 7b, 13b, 34b & $2023$  & \CheckmarkBold & Code-specialized \\
    CodeLlama-Python & 7b, 13b & $2023$  & \CheckmarkBold & Code-specialized \\
    StarCoder & 15b   & $2023$  & \CheckmarkBold & Code-specialized\\
    WizardCoder & 15b   & $2023$  & \CheckmarkBold 
 & Code-specialized \\
    \bottomrule
    \end{tabular}%
    }
  \caption{\textbf{Overview of the Evaluated Models}.}
  \label{tab:models_overview}%
\end{table}%
\section{The details of $23$ LLMs}
\label{sec:appendix_LLMs}
We have selected a total of $23$ mainstream LLMs, including both code-specialized models and general models, e.g., 

\noindent
\textbf{ChatGPT~\cite{NEURIPS2022_b1efde53, openai2023gpt4}:} ChatGPT was released by OpenAI in November $2022$ and has been widely recognized for its astonishing conversational generation capabilities. In March $2023$, OpenAI released ChatGPT $4.0$. In our experiments, we chose to use ChatGPT $3.5$ (gpt-3.5-turb) to explore its OOP.

\noindent
\textbf{GPT\_BigCode~\cite{allal2023santacoder}:} GPT\_BigCode, derived from the BigCode project, is a $1.12$ billion parameter model trained on subsets of Java, JavaScript, and Python from The Stack.

\noindent
\textbf{CodeLlama~\cite{roziere2023code}:} CodeLlama is a series of large-scale code language models based on Llama2 that offers state-of-the-art performance in open modeling, function completion, support for large input contexts, and zero-shot instruction following capabilities for programming tasks. CodeLlama includes the base model (CodeLlama), the Python specialized model (CodeLlama-Python), and the instruction-following model (CodeLlama-Instruct), each available with 7b, 13b, and 34b parameters. In our experiments, we selected the base models with 7b, 13b, and 34b parameters, as well as the Python-specialized models with 7b and 13b parameters.

\noindent
\textbf{WizardCoder~\cite{luo2023wizardcoder}:} WizardCoder is a model fine-tuned using the Evol-Instruct~\cite{xu2023wizardlm} method based on CodeLlama. WizardCoder includes the base model and the Python specialized model (WizardCoder-Python). The base model comes in 1b, 3b, and 15b variants, while the Python specialized model is available in 7b, 13b, and 34b. In our experiments, we selected the 15b version of the base model.

\noindent
\textbf{StarCoder~\cite{li2023starcoder}:} StarCoderBase is trained on The Stack (v1.2)~\footnote{\url{https://huggingface.co/datasets/bigcode/the-stack}} data in the GitHub repository. The StarCoder model is fine-tuned based on the StarCoderBase model.

\noindent
\textbf{Llama2~\cite{touvron2023llama}:} The Llama2 model was released by the Meta team in July 2023. Llama2 is a large language model (LLM) that has undergone pre-training and fine-tuning, with a range of parameters from 7 billion to 70 billion. In our experiments, we selected models with 7b, 13b, and 70b parameters.

\noindent
\textbf{InternLm~\cite{team2023internlm}:} InternLM encompasses models designed for practical scenarios. The InternLM model includes both a base model and a chat model with 7b and 20b parameters. In our experiments, we selected the base model with 7b parameters.

\noindent
\textbf{MPT~\cite{MosaicML2023Introducing}:} The MPT model is a decoder-style transformer trained by MosaicML. In our experiments, we selected the base model with 7b parameters.

\noindent
\textbf{DeepSeek~\cite{deepseekai2024deepseek}:} DeepSeek is an LLM based on the power-law scaling, encompassing models with 7b and 67b parameters. In our experiments, we opted to utilize the foundational models with 7b and 67b parameters.

\noindent
\textbf{Falcon~\cite{almazrouei2023falcon}:} The Falcon series models are primarily trained on diverse and high-quality corpora assembled from web data, including the 7b, 40b, and 180b parameter models. In our experiments, we opted to use models with 7b and 40b parameters.

\noindent
\textbf{Qwen~\cite{bai2023qwen}:} The Qwen model is a large language model based on the Transformer architecture, trained on a vast and diverse dataset for pre-training. The dataset encompasses a wide range of types, including extensive web text, professional books, code, and more. During our experiments, we selected the base models with 7b, 14b, and 72b parameters.

\noindent
\textbf{Yi~\footnote{\url{https://01.ai/cn}}:} The Yi series models are developed as bilingual language models with a focus on Chinese and English. Yi models are trained on a 3T multilingual corpus and demonstrate promising prospects in language understanding, common sense reasoning, and reading comprehension. In our experiments, we selected models with 6 billion and 34 billion parameters.

We use $23$ mainstream code-specialized and general models with the aim of better illustrating the performance of existing LLMs in OOP. The overview of the evaluated models is presented in Table~\ref{tab:models_overview}.


\section{Analysis of results}
\label{sec:appendix_Analysis_results}

In simple-level OOP of Table~\ref{tab:simple_level_OOP_scores}, ChatGPT scored $37.34$ at \textit{pass@$1$}. However, in the difficult-level and Moderate-level OOP, ChatGPT scored only $19.70$ and $2.53$ at \textit{pass@$1$}, respectively. CodeLlama-13b scored $16.21$ at \textit{pass@$1$} in the simple-level OOP. In the difficult-level and Moderate-level OOP, CodeLlama-13b scored only $0.00$ and $0.00$ at \textit{pass@$1$}, respectively. Additionally, WizardCoder-15b scored $16.79$ at \textit{pass@$1$} in the simple-level OOP., while in the difficult-level and Moderate-level OOP, it scored only $0.00$ and $0.00$ at \textit{pass@$1$}, respectively. It indicates that LLMs can comprehend and execute simple class, and public functions. However, their understanding of private functions, inheritance, and polymorphism is relatively weak.
It also provides us with room for improvement.

\section{Detailed description of the retrieval process}
\label{sec:retrieval_process}
During the retrieval process, we first search for the class \underline{\textbf{class}} and attribute variables \underline{\textbf{def \_\_init\_\_}}. Subsequently, we replace \underline{\textbf{def \_\_init\_\_}} in the generated code snippets with \textcolor{red}{\textbf{<endoftext>}}, and finally, we search for private functions \underline{\textbf{def \_}} and \underline{\textbf{def \_\_}}. Using this approach helps prevent the inadvertent retrieval of attribute variables as private functions during the search for private functions. The process of searching for public functions \underline{\textbf{def}} follows a similar method.

\begin{table*}[!t]
  \centering
  \resizebox{\linewidth}{!}{
    \begin{tabular}{cccccccccc}
    \toprule
    Model & \multicolumn{3}{c}{CodeLlama\_13b} & \multicolumn{3}{c}{WizardCoder\_15b} & \multicolumn{3}{c}{StarCoder} \\
    \midrule
    \textit{pass@$o$} & $1$     & $80$    & $100$   & $1$     & $80$    & $100$   & $1$     & $80$    & $100$ \\
    \midrule
    zero-shot CoT & $1.33_{\textcolor{red}{\textbf{(-1.59)}}}$  & $13.31_{\textcolor{blue}{\textbf{(+1.11)}}}$ & $14.62_{\textcolor{blue}{\textbf{(+1.51)}}}$ & $2.67_{\textcolor{red}{\textbf{(-0.35)}}}$  & $13.33_{\textcolor{red}{\textbf{(-3.89)}}}$ & $14.19_{\textcolor{red}{\textbf{(-4.18)}}}$ & $0.28_{\textcolor{red}{\textbf{(-0.98)}}}$  & $6.58_{\textcolor{red}{\textbf{(-3.47)}}}$  & $7.07_{\textcolor{red}{\textbf{(-3.81)}}}$ \\
    few-shot & $14.50_{\textcolor{blue}{\textbf{(+11.58)}}}$  & $48.13_{\textcolor{blue}{\textbf{(+35.93)}}}$ & $49.85_{\textcolor{blue}{\textbf{(+36.74)}}}$ & $17.34_{\textcolor{blue}{\textbf{(+14.32)}}}$ & $48.25_{\textcolor{blue}{\textbf{(+38.75)}}}$ & $49.78_{\textcolor{blue}{\textbf{(+39.77)}}}$ & $14.47_{\textcolor{blue}{\textbf{(+13.21)}}}$ & $46.59_{\textcolor{blue}{\textbf{(+36.54)}}}$ & $48.19_{\textcolor{blue}{\textbf{(+37.31)}}}$ \\
    few-shot CoT & $11.06_{\textcolor{red}{\textbf{(-3.44)}}}$ & $42.30_{\textcolor{red}{\textbf{(-5.83)}}}$  & $43.79_{\textcolor{red}{\textbf{(-6.06)}}}$ & $2.91_{\textcolor{red}{\textbf{(-14.43)}}}$  & $36.40_{\textcolor{red}{\textbf{(-11.85)}}}$  & $38.61_{\textcolor{red}{\textbf{(-11.17)}}}$ & $6.51_{\textcolor{red}{\textbf{(-7.96)}}}$  & $39.71_{\textcolor{red}{\textbf{(-6.88)}}}$ & $41.76_{\textcolor{red}{\textbf{(-6.43)}}}$ \\
    \bottomrule
    \end{tabular}%
    }
  \caption{\textbf{Performance of the CodeLlama\_13b, StarCoder, and WizardCoder\_15b models with advanced prompting strategies}, i.e., few-shot, zero-shot CoT, few-shot CoT, on the OOP benchmark. Additionally, we reported the delta in results between few-shot and few-shot CoT, zero-shot and zero-shot CoT, as well as between few-shot and zero-shot prompting strategies. (\sethlcolor{red}\hl{Red} indicates decline, while \sethlcolor{blue}\hl{blue} indicates increase.)}
  \label{tab:cot_methods}%
\end{table*}%

\section{Details of using the CoT strategy.}
\label{sec:few_cot}

\textbf{zero-shot CoT.} We incorporate "Let's think step by step" on top of the zero-shot, enabling LLMs to stepwise infer and thus complete the entire code generation process, as shown in Figure~\ref{fig:zero_shot_cot}.

\noindent
\textbf{few-shot.} We randomly selected three samples from MBPP~\cite{austin2021program}, but these three samples are limited to functions and do not involve relevant concepts and features of OOP. Subsequently, we manually re-write the selected three samples into examples of OOP based on the five major principles. Finally, the constructed samples were integrated into zero-shot to form a few-shot, as shown in Figure~\ref{fig:few_shot}.

\noindent
\textbf{few-shot CoT.} On the foundation of a few-shot, we first instruct the LLMs to generate corresponding steps based on the question and then proceed step by step to complete the entire code generation process, as shown in Figure~\ref{fig:few_shot_shot}.



\end{document}